\documentclass[11pt,a4paper]{article}

\usepackage[utf8]{inputenc}
\usepackage[T1]{fontenc}
\usepackage{graphicx}
\usepackage{amsmath}
\usepackage{amssymb}
\usepackage{booktabs}
\usepackage{multirow}
\usepackage{hyperref}
\usepackage{xcolor}
\usepackage{tikz}
\usepackage{algorithm}
\usepackage{algpseudocode}
\usepackage{listings}
\usepackage{geometry}
\usepackage{natbib}
\usepackage{subcaption}
\usepackage{pifont}
\usepackage{float}

\usetikzlibrary{shapes.geometric, arrows.meta, positioning, fit, backgrounds, calc, shapes.misc}

\tikzset{
    proarrow/.style={->, >=Stealth, line width=1.2pt, black!85},
    prodasharrow/.style={->, >=Stealth, dashed, line width=1.0pt, black!60},
    probox/.style={rectangle, draw=black!80, rounded corners=4pt, minimum width=2.2cm, minimum height=0.8cm, align=center, font=\small\sffamily, line width=1pt, fill=white},
    procontainer/.style={rectangle, draw=black!50, rounded corners=8pt, inner sep=12pt, line width=1pt},
    phaselabel/.style={font=\footnotesize\bfseries\sffamily, text=black!75},
    prodb/.style={cylinder, draw=black!80, shape border rotate=90, aspect=0.25, minimum height=1cm, minimum width=1.6cm, align=center, font=\small\sffamily, line width=1pt},
    stagebox/.style={rectangle, draw=black!50, dashed, rounded corners=6pt, inner sep=10pt, line width=1pt},
    infolabel/.style={font=\scriptsize\sffamily, text=black!55}
}

\definecolor{apiblue}{HTML}{4A90D9}
\definecolor{processgreen}{HTML}{27AE60}
\definecolor{mongoorange}{HTML}{E67E22}
\definecolor{vdbpurple}{HTML}{9B59B6}
\definecolor{fusionred}{HTML}{E74C3C}
\definecolor{inputgray}{HTML}{95A5A6}
\definecolor{llmyellow}{HTML}{F1C40F}
\definecolor{boostcyan}{HTML}{1ABC9C}

\newcommand{\fieldlabel}[1]{{\footnotesize\ttfamily\color{black!80}#1}}

\begin{document}

\begin{center}
\rule{0.4\textwidth}{0.4pt}
\end{center}

\vspace{2em}

\begin{flushleft}
{\fontsize{18}{22}\selectfont \textsc{Cognis: Context-Aware Memory for}\\[0.2em]
\textsc{Conversational AI Agents}}
\end{flushleft}

\vspace{0.8em}

\begin{flushleft}
\textbf{Parshva Daftari*}\hspace{0.8em}\textbf{Khush Patel*}\hspace{0.8em}\textbf{Shreyas Kapale*}\hspace{0.8em}\textbf{Jithin George}\hspace{0.8em}\textbf{Siva Surendira}\\[0.3em]
Lyzr Research\\[0.1em]
\texttt{\{parshva, khush, shreyas, jithin, siva\}@lyzr.ai}
\renewcommand{\thefootnote}{\fnsymbol{footnote}}
\footnotetext[1]{Equal contribution}
\end{flushleft}

\vspace{1.5em}

\begin{abstract}
LLM agents lack persistent memory, causing conversations to reset each session and preventing personalization over time. We present \textbf{Lyzr Cognis}, a unified memory architecture for conversational AI agents that addresses this limitation through a multi-stage retrieval pipeline. Cognis combines a dual-store backend pairing OpenSearch BM25 keyword matching with Matryoshka vector similarity search, fused via Reciprocal Rank Fusion. Its context-aware ingestion pipeline retrieves existing memories before extraction, enabling intelligent version tracking that preserves full memory history while keeping the store consistent. Temporal boosting enhances time-sensitive queries, and a BGE-2 cross-encoder reranker refines final result quality. We evaluate Cognis on two independent benchmarks---LoCoMo and LongMemEval---across eight answer generation models, demonstrating state-of-the-art performance on both. The system is open-source and deployed in production serving conversational AI applications.
\end{abstract}

\section{Introduction}

The rapid advancement of Large Language Models (LLMs) has enabled the development of sophisticated conversational AI agents capable of complex reasoning and natural language understanding. However, these agents face a fundamental limitation: they operate within fixed context windows and lack persistent memory capabilities, causing each conversation to begin without knowledge of prior interactions.

This limitation manifests in several practical problems:
\begin{itemize}
    \item \textbf{Conversation discontinuity}: Users must re-establish context in every session
    \item \textbf{Lost personalization}: Agents cannot learn user preferences over time
    \item \textbf{Repetitive interactions}: Users repeatedly provide the same information
    \item \textbf{Shallow relationships}: Agents cannot build rapport or trust through continuity
\end{itemize}

Existing approaches to LLM memory fall into two categories: \textit{retrieval-augmented generation} (RAG), which treats memory as a document retrieval problem, and \textit{specialized memory systems} like Mem0~\citep{mem0}, Zep~\citep{zep}, and SuperMemory~\citep{supermemory}, which provide memory-specific abstractions. While these systems represent important progress, they often rely on single retrieval modalities, lack sophisticated temporal reasoning, and do not maintain version history for evolving information.

We present \textbf{Lyzr Cognis}, a unified memory architecture designed to address these limitations. Our key contributions are:

\begin{enumerate}
    \item \textbf{Memory Taxonomy}: A comprehensive classification system with 15 semantic categories (e.g., personal details, professional, health) and 2 persistence scopes (USER for cross-session, CONTEXT for session-specific).

    \item \textbf{Dual-Store Architecture}: A streamlined storage layer combining:
    \begin{itemize}
        \item OpenSearch for document storage, native BM25 search with configurable text analysis, and version history
        \item Vector Database (VDB) for dual-dimension Matryoshka embeddings (768D + 256D) enabling efficient two-stage semantic search
    \end{itemize}

    \item \textbf{Context-Aware Ingestion}: An intelligent extraction pipeline that retrieves similar existing memories from the VDB \textit{before} LLM processing, enabling the model to make informed decisions about whether to ADD new facts, UPDATE existing ones (with version linking), DELETE contradicted information, or skip duplicates entirely.

    \item \textbf{Hybrid Retrieval Pipeline}: A sophisticated search system combining vector similarity and BM25 keyword matching through Reciprocal Rank Fusion (RRF) with 70\% vector and 30\% BM25 weighting, temporal boosting for time-aware queries, content deduplication, and a BGE-2 cross-encoder reranker for final result refinement.

    \item \textbf{Version Tracking}: Full version history with \texttt{is\_current} flags and \texttt{replaces\_id} links, enabling historical queries like ``What were all my previous jobs?''

    \item \textbf{Cross-Benchmark Validation}: Comprehensive evaluation on both LoCoMo and LongMemEval benchmarks across eight answer generation models, demonstrating that architectural advantages generalize across different evaluation frameworks and LLM backends, with up to 92.4\% accuracy on LongMemEval's 500-question benchmark.
\end{enumerate}

\begin{figure}[h]
\centering
\resizebox{0.95\textwidth}{!}{%
\begin{tikzpicture}[
    bar/.style={draw, minimum width=0.6cm, anchor=south},
    label/.style={font=\scriptsize}
]

\definecolor{mem0color}{HTML}{5DADE2}
\definecolor{zepcolor}{HTML}{48C9B0}
\definecolor{mem0gcolor}{HTML}{F4D03F}
\definecolor{lyzrcolor}{HTML}{E67E22}

\def\singlehop{0}
\def\multihop{3.5}
\def\opendomain{7}
\def\temporal{10.5}

\def\scale{0.055}

\node[label, anchor=north] at (\singlehop+0.95, -0.3) {Single-Hop};
\fill[mem0color] (\singlehop, 0) rectangle (\singlehop+0.4, 38.72*\scale);
\fill[zepcolor] (\singlehop+0.5, 0) rectangle (\singlehop+0.9, 35.74*\scale);
\fill[mem0gcolor] (\singlehop+1.0, 0) rectangle (\singlehop+1.4, 38.09*\scale);
\fill[lyzrcolor] (\singlehop+1.5, 0) rectangle (\singlehop+1.9, 48.66*\scale);
\node[label, font=\scriptsize\bfseries, text=lyzrcolor] at (\singlehop+1.7, 48.66*\scale+0.2) {+25.7\%};

\node[label, anchor=north] at (\multihop+0.95, -0.3) {Multi-Hop};
\fill[mem0color] (\multihop, 0) rectangle (\multihop+0.4, 28.64*\scale);
\fill[zepcolor] (\multihop+0.5, 0) rectangle (\multihop+0.9, 19.37*\scale);
\fill[mem0gcolor] (\multihop+1.0, 0) rectangle (\multihop+1.4, 24.32*\scale);
\fill[lyzrcolor] (\multihop+1.5, 0) rectangle (\multihop+1.9, 31.51*\scale);
\node[label, font=\scriptsize\bfseries, text=lyzrcolor] at (\multihop+1.7, 31.51*\scale+0.2) {+10.0\%};

\node[label, anchor=north] at (\opendomain+0.95, -0.3) {Open-Domain};
\fill[mem0color] (\opendomain, 0) rectangle (\opendomain+0.4, 47.65*\scale);
\fill[zepcolor] (\opendomain+0.5, 0) rectangle (\opendomain+0.9, 49.56*\scale);
\fill[mem0gcolor] (\opendomain+1.0, 0) rectangle (\opendomain+1.4, 49.27*\scale);
\fill[lyzrcolor] (\opendomain+1.5, 0) rectangle (\opendomain+1.9, 54.77*\scale);
\node[label, font=\scriptsize\bfseries, text=lyzrcolor] at (\opendomain+1.7, 54.77*\scale+0.2) {+10.5\%};

\node[label, anchor=north] at (\temporal+0.95, -0.3) {Temporal};
\fill[mem0color] (\temporal, 0) rectangle (\temporal+0.4, 48.93*\scale);
\fill[zepcolor] (\temporal+0.5, 0) rectangle (\temporal+0.9, 42.00*\scale);
\fill[mem0gcolor] (\temporal+1.0, 0) rectangle (\temporal+1.4, 51.55*\scale);
\fill[lyzrcolor] (\temporal+1.5, 0) rectangle (\temporal+1.9, 62.68*\scale);
\node[label, font=\scriptsize\bfseries, text=lyzrcolor] at (\temporal+1.7, 62.68*\scale+0.2) {+21.6\%};

\draw[->] (-0.5, 0) -- (-0.5, 3.9);
\node[label, rotate=90, anchor=south] at (-1.3, 1.8) {F1 Score};
\foreach \y in {0, 20, 40, 60} {
    \draw (-0.6, \y*\scale) -- (-0.5, \y*\scale);
    \node[label, anchor=east] at (-0.7, \y*\scale) {\y};
}

\draw (-0.5, 0) -- (12.9, 0);

\node[label] at (14, 3.5) {\textbf{Legend}};
\fill[mem0color] (13.2, 3) rectangle (13.7, 3.2);
\node[label, anchor=west] at (13.8, 3.1) {Mem0};
\fill[zepcolor] (13.2, 2.5) rectangle (13.7, 2.7);
\node[label, anchor=west] at (13.8, 2.6) {Zep};
\fill[mem0gcolor] (13.2, 2) rectangle (13.7, 2.2);
\node[label, anchor=west] at (13.8, 2.1) {Mem0g};
\fill[lyzrcolor] (13.2, 1.5) rectangle (13.7, 1.7);
\node[label, anchor=west] at (13.8, 1.6) {\textbf{Cognis}};

\end{tikzpicture}%
}
\caption{LoCoMo benchmark F1 scores across four question types. Cognis achieves the highest F1 in every category. Percentage labels show Cognis's gain over the strongest baseline. Mem0 = Mem0 retrieval; Zep = Zep memory framework; Mem0g = Mem0 with graph memory enabled; Cognis = Lyzr Cognis (ours).}
\label{fig:hero}
\end{figure}
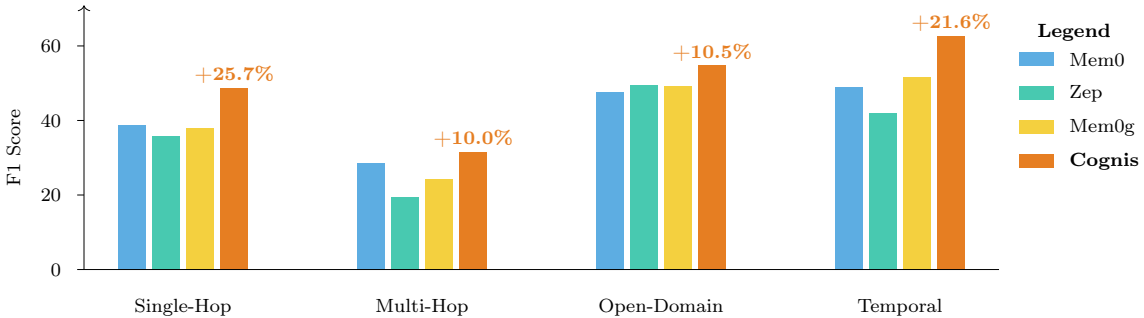

\begin{figure}[h]
\centering
\resizebox{0.95\textwidth}{!}{%
\begin{tikzpicture}[
    bar/.style={draw, minimum width=0.6cm, anchor=south},
    label/.style={font=\scriptsize}
]

\definecolor{zepgcolor}{HTML}{48C9B0}
\definecolor{supercolor}{HTML}{9B59B6}
\definecolor{cogniscolor}{HTML}{E67E22}

\def\ssuser{0}
\def\ssasst{2.5}
\def\sspref{5.0}
\def\knowupd{7.5}
\def\temporal{10.0}
\def\multisess{12.5}

\def\scale{0.045}

\node[label, anchor=north] at (\ssuser+0.65, -0.3) {\shortstack{SS-\\User}};
\fill[zepgcolor] (\ssuser, 0) rectangle (\ssuser+0.35, 92.9*\scale);
\fill[supercolor] (\ssuser+0.45, 0) rectangle (\ssuser+0.8, 97.1*\scale);
\fill[cogniscolor] (\ssuser+0.9, 0) rectangle (\ssuser+1.25, 100.0*\scale);
\node[label, font=\scriptsize\bfseries, text=cogniscolor] at (\ssuser+1.1, 100.0*\scale+0.15) {+3.0\%};

\node[label, anchor=north] at (\ssasst+0.65, -0.3) {\shortstack{SS-\\Asst}};
\fill[zepgcolor] (\ssasst, 0) rectangle (\ssasst+0.35, 80.4*\scale);
\fill[supercolor] (\ssasst+0.45, 0) rectangle (\ssasst+0.8, 96.4*\scale);
\fill[cogniscolor] (\ssasst+0.9, 0) rectangle (\ssasst+1.25, 92.9*\scale);

\node[label, anchor=north] at (\sspref+0.65, -0.3) {\shortstack{SS-\\Pref}};
\fill[zepgcolor] (\sspref, 0) rectangle (\sspref+0.35, 56.7*\scale);
\fill[supercolor] (\sspref+0.45, 0) rectangle (\sspref+0.8, 70.0*\scale);
\fill[cogniscolor] (\sspref+0.9, 0) rectangle (\sspref+1.25, 93.3*\scale);
\node[label, font=\scriptsize\bfseries, text=cogniscolor] at (\sspref+1.1, 93.3*\scale+0.15) {+33.3\%};

\node[label, anchor=north] at (\knowupd+0.65, -0.3) {\shortstack{Know.\\Update}};
\fill[zepgcolor] (\knowupd, 0) rectangle (\knowupd+0.35, 83.3*\scale);
\fill[supercolor] (\knowupd+0.45, 0) rectangle (\knowupd+0.8, 88.5*\scale);
\fill[cogniscolor] (\knowupd+0.9, 0) rectangle (\knowupd+1.25, 96.2*\scale);
\node[label, font=\scriptsize\bfseries, text=cogniscolor] at (\knowupd+1.1, 96.2*\scale+0.15) {+8.7\%};

\node[label, anchor=north] at (\temporal+0.65, -0.3) {\shortstack{Temp.\\Reason.}};
\fill[zepgcolor] (\temporal, 0) rectangle (\temporal+0.35, 62.4*\scale);
\fill[supercolor] (\temporal+0.45, 0) rectangle (\temporal+0.8, 76.7*\scale);
\fill[cogniscolor] (\temporal+0.9, 0) rectangle (\temporal+1.25, 92.5*\scale);
\node[label, font=\scriptsize\bfseries, text=cogniscolor] at (\temporal+1.1, 92.5*\scale+0.15) {+20.6\%};

\node[label, anchor=north] at (\multisess+0.65, -0.3) {\shortstack{Multi-\\Session}};
\fill[zepgcolor] (\multisess, 0) rectangle (\multisess+0.35, 57.9*\scale);
\fill[supercolor] (\multisess+0.45, 0) rectangle (\multisess+0.8, 71.4*\scale);
\fill[cogniscolor] (\multisess+0.9, 0) rectangle (\multisess+1.25, 87.2*\scale);
\node[label, font=\scriptsize\bfseries, text=cogniscolor] at (\multisess+1.1, 87.2*\scale+0.15) {+22.1\%};

\draw[->] (-0.5, 0) -- (-0.5, 4.8);
\node[label, rotate=90, anchor=south] at (-1.3, 2.2) {Accuracy (\%)};
\foreach \y in {0, 20, 40, 60, 80, 100} {
    \draw (-0.6, \y*\scale) -- (-0.5, \y*\scale);
    \node[label, anchor=east] at (-0.7, \y*\scale) {\y};
}

\draw (-0.5, 0) -- (14.2, 0);

\node[label] at (15.2, 4.3) {\textbf{Legend}};
\fill[zepgcolor] (14.4, 3.8) rectangle (14.9, 4.0);
\node[label, anchor=west] at (15.0, 3.9) {Zep/Graphiti};
\fill[supercolor] (14.4, 3.3) rectangle (14.9, 3.5);
\node[label, anchor=west] at (15.0, 3.4) {SuperMemory};
\fill[cogniscolor] (14.4, 2.8) rectangle (14.9, 3.0);
\node[label, anchor=west] at (15.0, 2.9) {\textbf{Cognis}};

\end{tikzpicture}%
}
\caption{Cross-system accuracy on LongMemEval across six question types. Cognis (orange) leads on five of six categories, with the largest gains on preference recall (+33.3\%) and temporal reasoning (+20.6\%). SuperMemory leads only on SS-Assistant. Zep/Graphiti = Zep with Graphiti; SuperMemory = SuperMemory graph memory; Cognis = Lyzr Cognis (ours, best across answer models).}
\label{fig:longmemeval-hero}
\end{figure}
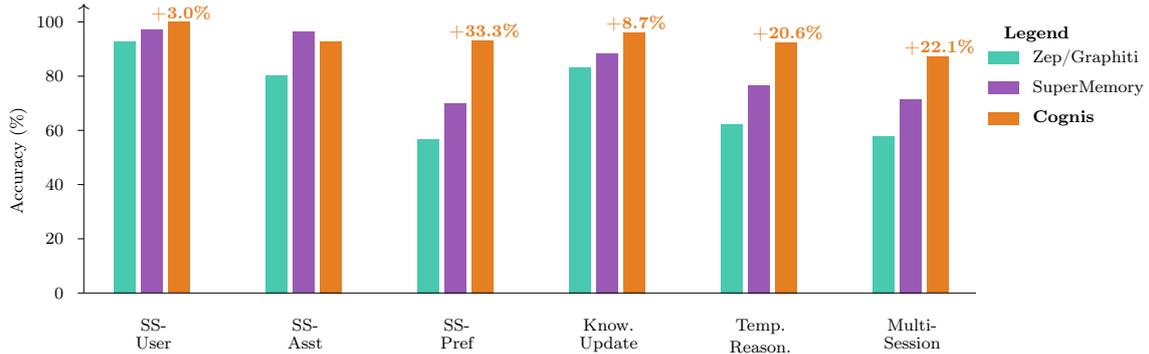

The remainder of this paper is organized as follows: Section~\ref{sec:related} reviews related work, Section~\ref{sec:architecture} describes the system architecture, Section~\ref{sec:ingestion} details the ingestion pipeline, Section~\ref{sec:retrieval} explains the retrieval pipeline, Section~\ref{sec:setup} describes experimental setup, Section~\ref{sec:locomo-results} presents results on the LoCoMo benchmark, Section~\ref{sec:longmemeval-results} presents results on LongMemEval, Section~\ref{sec:discussion} discusses findings and limitations, and Section~\ref{sec:conclusion} concludes.

\section{Related Work}
\label{sec:related}

\subsection{Memory Systems for LLM Agents}

The challenge of providing persistent memory to LLM agents has spawned numerous approaches. Early commercial solutions include \textbf{Mem0}~\citep{mem0}, which provides a memory layer with automatic fact extraction and vector-based retrieval, and \textbf{Zep}~\citep{zep}, which offers long-term memory with session management and temporal awareness. \textbf{SuperMemory}~\citep{supermemory} focuses on knowledge graph integration for multi-hop reasoning capabilities.

Recent academic work has advanced agent memory architectures significantly. \textbf{MemGPT}~\citep{memgpt} reconceptualizes memory management through an operating system lens, treating the LLM as a processor with explicit memory hierarchies---main context as RAM and external storage as disk. This approach enables virtual context management beyond fixed window limits but requires complex memory paging operations. Unlike MemGPT's OS-style paging between context (RAM) and storage (disk), which requires explicit memory management operations to decide what to swap in and out, our dual-store architecture keeps all memories simultaneously accessible through parallel retrieval---sacrificing the theoretical elegance of hierarchical memory for practical simplicity and lower operational complexity. \textbf{MemoryBank}~\citep{memorybank} enhances LLMs with long-term memory through memory consolidation mechanisms inspired by human cognition, storing and updating memories during conversations and employing a memory retrieval mechanism during inference.

\textbf{ReadAgent}~\citep{readagent} takes a human-inspired approach to processing long documents, building ``gist memory'' that captures essential information at multiple granularities. The agent learns to decide what to remember and what to retrieve, mimicking how humans selectively process and retain information from lengthy texts. While ReadAgent focuses on reading comprehension of long documents, our system addresses a different challenge: maintaining coherent memory across many short conversational exchanges over extended time periods, where the primary difficulty is not document length but temporal span and information evolution. \textbf{SimpleMem}~\citep{simplemem} proposes efficient lifelong memory through practical mechanisms for memory organization and retrieval, focusing on computational efficiency without sacrificing effectiveness.

\textbf{A-MEM}~\citep{amem} introduces agentic memory mechanisms with structured approaches to memory management, allowing agents to autonomously organize and access their memory stores. \textbf{MemR$^3$}~\citep{memr3} combines reflective reasoning with memory retrieval, enabling agents to assess memory relevance through reasoning rather than simple similarity matching---a departure from purely embedding-based approaches. Our context-aware ingestion shares this philosophy of using reasoning over raw similarity, but applies it at write time (deciding how new information relates to existing memories) rather than read time (deciding which memories are relevant to a query). The \textbf{Hindsight Memory} approach~\citep{hindsight} identifies three core capabilities essential for effective agent memory: retention (what to store), recall (how to retrieve), and reflection (how to learn from past experiences) for improved decision-making.

Our work synthesizes insights from these systems while addressing their limitations: (1) unlike MemGPT's complex paging, we use a simpler dual-store architecture with hybrid retrieval; (2) unlike single-modality systems, we combine multiple retrieval approaches through RRF fusion; (3) we implement Matryoshka embeddings~\citep{matryoshka} for efficient two-stage retrieval; (4) we maintain full version history for evolving information; and (5) we provide comprehensive temporal reasoning often lacking in existing systems.

\subsection{Retrieval-Augmented Generation}

RAG systems~\citep{lewis2020rag} augment LLM responses with retrieved context from external knowledge bases, achieving strong performance on knowledge-intensive tasks without parameter updates. While effective for static document retrieval, standard RAG architectures lack the temporal awareness, version tracking, and memory update mechanisms needed for conversational memory where information evolves over time.

Recent advances have improved RAG capabilities. \textbf{CLaRa}~\citep{clara} bridges retrieval and generation through continuous latent reasoning, enabling joint optimization of both processes rather than treating them as separate stages. This unified approach allows the model to reason over retrieved information more effectively. Our hybrid retrieval approach extends RAG principles with BM25 keyword matching for exact term handling, temporal boosting for time-aware queries, and BGE-2~\citep{bge2024} cross-encoder reranking for final result refinement.

\subsection{Hybrid Search and Dense Retrieval}

The complementary strengths of dense (embedding-based) and sparse (keyword-based) retrieval have motivated hybrid approaches~\citep{ma2021hybrid}. Dense retrieval excels at semantic similarity and paraphrase matching, while sparse methods like BM25~\citep{bm25} capture exact term matches that embeddings may miss. Reciprocal Rank Fusion (RRF)~\citep{cormack2009rrf} provides a simple yet effective method for combining ranked lists from multiple retrievers without requiring score calibration.

The accuracy benefits of hybrid search are well-documented. Queries containing specific names, dates, or technical terms are often missed by embedding-based retrieval alone because dense models collapse lexically distinct but semantically similar tokens into nearby vectors. BM25 anchors on exact tokens and recovers these cases, while vector search handles paraphrased or semantically equivalent queries that keyword matching would miss. By fusing both ranked lists through RRF, the resulting pipeline achieves higher recall and precision than either modality in isolation, particularly on entity-heavy and temporal queries where one modality alone consistently underperforms.

Recent work on dense retrieval optimization includes \textbf{CADET}~\citep{cadet}, which demonstrates that cross-encoder listwise distillation with synthetic data can significantly improve dense retriever performance beyond conventional contrastive learning approaches. This finding informed our decision to include a BGE-2 cross-encoder reranker as a final refinement stage. Adding a cross-encoder after initial bi-encoder retrieval provides a second pass of fine-grained relevance scoring, catching subtle mismatches that bi-encoder dot-product similarity overlooks and yielding measurable accuracy gains on multi-hop and open-domain queries. We extend these techniques with explicit temporal relevance scoring to address a gap in existing hybrid search systems.

\subsection{Embedding Representations}

Modern embedding models have achieved strong performance on semantic similarity tasks. \textbf{Matryoshka Representation Learning}~\citep{matryoshka} introduced the concept of training embeddings that remain effective when truncated to lower dimensions, enabling adaptive compute-accuracy tradeoffs. We leverage this property for two-stage retrieval: fast shortlisting with truncated 256D embeddings followed by accurate ranking with full 768D embeddings.

The \textbf{BGE M3-Embedding}~\citep{bge2024} family provides multi-lingual, multi-functionality embeddings through self-knowledge distillation. The associated BGE reranker models serve as cross-encoders that jointly encode query-document pairs, providing more accurate relevance judgments than bi-encoder similarity at the cost of increased computation. We use the BGE-2 reranker as our final refinement stage.

\subsection{Query Understanding and Attention}

Effective memory retrieval depends on understanding user intent. \textbf{System 2 Attention}~\citep{s2a} addresses attention limitations in Transformers by using LLM reasoning to filter irrelevant context before processing, improving focus on pertinent information. \textbf{Rephrase and Respond}~\citep{rar} shows that having LLMs reformulate questions before answering improves performance across diverse tasks by clarifying ambiguous queries. These insights inform our query analysis pipeline, which detects temporal intent (triggering time-based boosting) and history keywords (enabling version chain traversal).

\subsection{Cognitive Science Foundations}

Our memory taxonomy draws on cognitive science research distinguishing memory types. Tulving's foundational work~\citep{tulving1972episodic} established the distinction between episodic memory (autobiographical events) and semantic memory (factual knowledge), which informs our decay rate assignments. The Atkinson-Shiffrin model~\citep{atkinson1968human} of human memory with its multi-store architecture (sensory, short-term, long-term) inspired our separation of immediate recall (raw messages) from consolidated memories (extracted facts).

\subsection{Benchmarks for Long-Term Memory}

Evaluating long-term memory systems requires specialized benchmarks that test persistence and temporal reasoning. \textbf{LongMemEval}~\citep{longmemeval} provides a comprehensive benchmark evaluating five critical capabilities: information extraction, multi-session reasoning, temporal reasoning, knowledge updates, and abstention (correctly declining to answer when information is unavailable). The benchmark reveals a concerning 30\% accuracy drop in commercial systems during sustained interactions, highlighting the difficulty of maintaining coherent long-term memory.

The \textbf{LoCoMo} benchmark~\citep{locomo} specifically tests memory systems across multi-session conversations requiring recall across 50+ sessions, with distinct question categories (single-hop, multi-hop, open-domain, temporal) that expose different failure modes. We adopt LongMemEval's evaluation methodology with question type-specific judge prompts while evaluating on LoCoMo's challenging multi-session scenarios.

\section{System Architecture}
\label{sec:architecture}

\subsection{Overview}

The Lyzr Cognis system follows a streamlined architecture consisting of an orchestration engine (the Unified Memory Provider) and a dual-store storage backend. This design prioritizes simplicity and efficiency while maintaining powerful hybrid retrieval capabilities.

The architecture deliberately avoids unnecessary complexity. While graph databases and caching layers can provide benefits in specific scenarios, our empirical evaluation showed that the combination of OpenSearch's native BM25 search with dual-dimension vector search provides excellent retrieval quality with simpler operational requirements. Our ablation studies (Section~\ref{sec:ablation}) demonstrate that OpenSearch's native BM25 implementation significantly outperforms MongoDB's text indexing, particularly on open-domain queries requiring broad entity matching.

\subsection{Memory Taxonomy}
\label{sec:taxonomy}

We introduce a comprehensive taxonomy for classifying memories along three dimensions, drawing on cognitive science research into human memory organization~\citep{tulving1972episodic, atkinson1968human}.

\subsection{Dual-Store Architecture}

Our architecture employs two complementary storage systems, each optimized for different access patterns (Table~\ref{tab:stores}).

\begin{table}[h]
\centering
\caption{Dual-Store Data Distribution}
\label{tab:stores}
\begin{tabular}{llp{5.5cm}}
\toprule
\textbf{Store} & \textbf{Data Type} & \textbf{Purpose} \\
\midrule
OpenSearch & Documents & Primary store for memories and messages, native BM25 search with configurable text analysis, version history tracking \\
VDB & Vectors (768D + 256D) & Matryoshka semantic search with dual-dimension collections \\
\bottomrule
\end{tabular}
\end{table}

\subsubsection{OpenSearch Indexes}

OpenSearch serves as the document store with two primary indexes. The \textbf{messages} index stores raw conversation messages with timestamps, speaker information, and processing status flags. The \textbf{memories} index contains extracted facts with full metadata including category, scope, version tracking fields (\texttt{is\_current}, \texttt{replaces\_id}), and event timestamps for temporal reasoning.

OpenSearch provides native BM25 search with configurable text analyzers, offering superior scoring control compared to MongoDB's text indexing. Our ablation experiments show that this switch from MongoDB to OpenSearch yields the single largest performance improvement, particularly a +20.3\% gain on open-domain LLM Judge scores (Section~\ref{sec:ablation}). The native BM25 implementation with proper tokenization and term-frequency weighting excels at matching specific entity names, dates, and technical terms that pure semantic search may miss.

\subsubsection{VDB Collections}

The vector database maintains three collections for efficient retrieval:

\begin{table}[h]
\centering
\caption{VDB Collection Structure}
\label{tab:vdb-collections}
\begin{tabular}{lcp{5cm}}
\toprule
\textbf{Collection} & \textbf{Dimension} & \textbf{Purpose} \\
\midrule
\texttt{*\_768d} & 768D & Memory embeddings for accurate semantic retrieval \\
\texttt{*\_256d} & 256D & Memory embeddings for fast shortlisting (200 candidates) \\
\texttt{immediate\_recall\_256d} & 256D & Raw message embeddings for immediate recall queries \\
\bottomrule
\end{tabular}
\end{table}

\begin{figure}[h]
\centering
\resizebox{0.9\textwidth}{!}{%
\begin{tikzpicture}[
    archdb/.style={cylinder, draw=black!80, shape border rotate=90, aspect=0.25, minimum height=1.4cm, minimum width=3.5cm, align=center, line width=1.2pt, font=\large\sffamily\bfseries},
    collection/.style={rectangle, draw=black!60, rounded corners=5pt, minimum width=5.5cm, minimum height=1.0cm, align=center, font=\normalsize\sffamily, line width=1pt},
    detail/.style={font=\small\sffamily, text=black!55, align=center}
]

\node[procontainer, fill=mongoorange!10, minimum width=7.5cm, minimum height=7cm] (mongobox) at (0, 0) {};
\node[archdb, fill=mongoorange!35] at (0, 2.4) {OpenSearch};
\node[font=\normalsize\sffamily\bfseries, text=mongoorange!90] at (0, 1.4) {Native BM25};
\node[collection, fill=mongoorange!20] (um) at (0, 0.4) {memories};
\node[collection, fill=mongoorange!20] (sm) at (0, -0.8) {messages};
\node[detail] at (0, -2.0) {native BM25, version tracking};

\node[procontainer, fill=vdbpurple!10, minimum width=7.5cm, minimum height=7cm] (vdbbox) at (9.5, 0) {};
\node[archdb, fill=vdbpurple!35] at (9.5, 2.4) {Vector DB};
\node[collection, fill=vdbpurple!20] (v768) at (9.5, 0.8) {*\_768d (accurate)};
\node[collection, fill=vdbpurple!20] (v256) at (9.5, -0.3) {*\_256d (fast)};
\node[collection, fill=vdbpurple!20] (vimm) at (9.5, -1.4) {immediate\_recall};
\node[detail] at (9.5, -2.5) {Matryoshka embeddings};

\end{tikzpicture}%
}
\caption{Dual-store architecture: OpenSearch (documents + native BM25) and VDB (Matryoshka embeddings at 768D/256D).}
\label{fig:multi-store}
\end{figure}
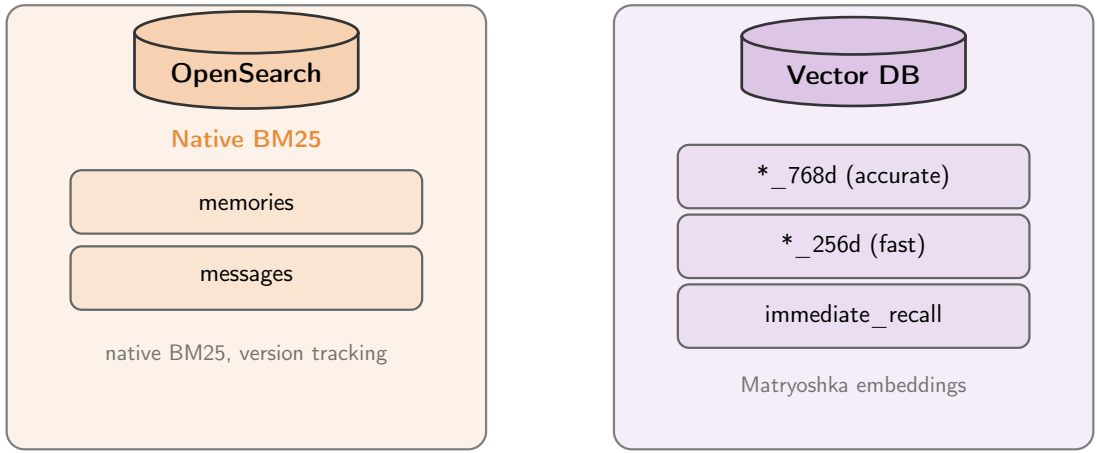

This dual-store approach provides the best of both worlds: OpenSearch's native BM25 search with configurable text analysis combined with the VDB's high-performance vector similarity search. The separation also allows independent scaling of document and vector workloads.

\section{Ingestion Pipeline}
\label{sec:ingestion}

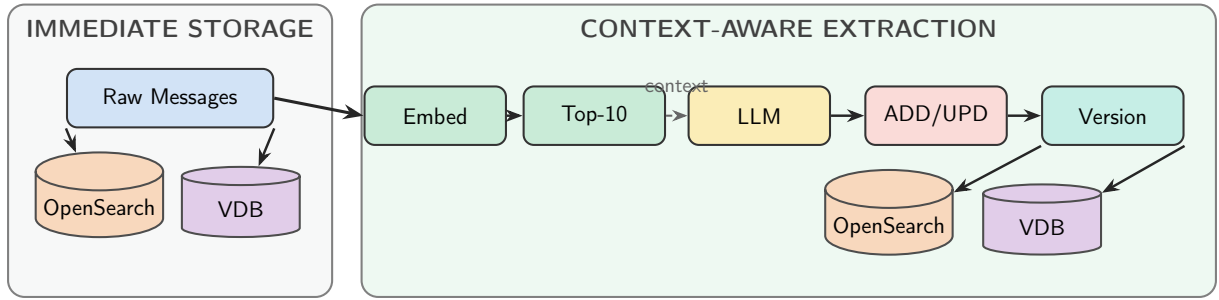
\begin{figure}[h]
\centering
\resizebox{\textwidth}{!}{%
\begin{tikzpicture}[
    node distance=1.0cm and 1.5cm,
    ingbox/.style={rectangle, draw=black!80, rounded corners=5pt, minimum width=2.4cm, minimum height=1.0cm, align=center, font=\normalsize\sffamily, line width=1pt},
    ingdb/.style={cylinder, draw=black!70, shape border rotate=90, aspect=0.3, minimum height=1.2cm, minimum width=2.0cm, align=center, font=\normalsize\sffamily, line width=1pt}
]

\node[procontainer, fill=inputgray!8, minimum width=5.5cm, minimum height=5.0cm] (leftpanel) at (0, 0) {};
\node[phaselabel, font=\large\bfseries\sffamily] at (0, 2.1) {IMMEDIATE STORAGE};

\node[ingbox, fill=apiblue!25, minimum width=3.5cm] (input) at (0, 0.9) {Raw Messages};
\node[ingdb, fill=mongoorange!30] (mongo1) at (-1.2, -1.0) {OpenSearch};
\node[ingdb, fill=vdbpurple!30] (vdb1) at (1.2, -1.0) {VDB};

\draw[proarrow] (input.south west) -- (mongo1);
\draw[proarrow] (input.south east) -- (vdb1);

\node[procontainer, fill=processgreen!8, minimum width=14.5cm, minimum height=5.0cm] (rightpanel) at (10.5, 0) {};
\node[phaselabel, font=\large\bfseries\sffamily] at (10.5, 2.1) {CONTEXT-AWARE EXTRACTION};

\node[ingbox, fill=processgreen!25] (embed) at (4.5, 0.6) {Embed};
\node[ingbox, fill=processgreen!25] (retrieve) at (7.2, 0.6) {Top-10};
\node[ingbox, fill=llmyellow!30] (llm) at (10.0, 0.6) {LLM};
\node[ingbox, fill=fusionred!20] (ops) at (13.0, 0.6) {ADD/UPD};
\node[ingbox, fill=boostcyan!25] (ver) at (16.0, 0.6) {Version};

\node[ingdb, fill=mongoorange!30] (mongo2) at (12.2, -1.3) {OpenSearch};
\node[ingdb, fill=vdbpurple!30] (vdb2) at (14.8, -1.3) {VDB};

\draw[proarrow] (embed) -- (retrieve);
\draw[prodasharrow] (retrieve) -- (llm);
\node[font=\small\sffamily, text=black!60] at (8.6, 1.1) {context};
\draw[proarrow] (llm) -- (ops);
\draw[proarrow] (ops) -- (ver);

\draw[proarrow] (ver.south west) -- (mongo2);
\draw[proarrow] (ver.south east) -- (vdb2);

\draw[proarrow, line width=1.5pt] (input.east) -- (embed.west);

\end{tikzpicture}%
}
\caption{Two-panel ingestion pipeline. \textbf{Left}: Immediate storage enables recall before extraction. \textbf{Right}: Context-aware extraction retrieves similar memories, LLM decides ADD/UPDATE/DELETE, version tracking maintains history.}
\label{fig:ingestion}
\end{figure}

The ingestion pipeline transforms raw conversation messages into structured, searchable memories. A key innovation is that the LLM extraction step receives context from similar existing memories, enabling intelligent decisions about how to handle new information relative to what the system already knows (Figure~\ref{fig:ingestion}).

\subsection{Message Storage and Immediate Recall}

When messages arrive via the API, they undergo two immediate storage operations. First, OpenSearch stores messages in the \texttt{messages} index with metadata including timestamp, speaker, session ID, and a \texttt{processed: false} flag. Simultaneously, a 256D embedding is generated and stored in the VDB's \texttt{immediate\_recall\_256d} collection, enabling instant retrieval of recent conversation context. This dual storage ensures that even before fact extraction completes, the system can retrieve relevant conversation history for queries requiring immediate recall.

\subsection{Speaker Identification}

Messages follow a structured format that enables speaker identification:

\begin{lstlisting}
[2024-05-08 10:30:00] James: I just got a new job at Google!
[2024-05-08 10:30:15] Assistant: Congratulations! That's exciting news.
\end{lstlisting}

The extraction system parses this format to extract the speaker name (``James'') for proper attribution, calculate absolute dates from relative references (``yesterday'' $\rightarrow$ specific date based on message timestamp), and associate facts with the correct user in multi-party conversations.

\subsection{Context-Aware LLM Extraction}

\begin{figure}[h]
\centering
\resizebox{\textwidth}{!}{%
\begin{tikzpicture}[
    node distance=1.2cm,
    cmpbox/.style={rectangle, draw=black!80, rounded corners=5pt, minimum width=7.0cm, minimum height=1.0cm, align=center, font=\normalsize\sffamily, line width=1pt},
    badresult/.style={rectangle, draw=fusionred!80, fill=fusionred!12, rounded corners=5pt, minimum width=7.0cm, minimum height=1.8cm, align=left, font=\normalsize\sffamily, line width=1.2pt},
    goodresult/.style={rectangle, draw=processgreen!80, fill=processgreen!12, rounded corners=5pt, minimum width=7.0cm, minimum height=1.8cm, align=left, font=\normalsize\sffamily, line width=1.2pt},
    title/.style={font=\large\bfseries\sffamily, text=black!80}
]

\node[title] at (-4.5, 5.5) {Without Context Retrieval};

\node[cmpbox, fill=inputgray!15] (msg1) at (-4.5, 4.3) {New: ``Promoted to Sr. Engineer''};
\node[cmpbox, fill=inputgray!8, draw=black!35, text=black!45] (exist1) at (-4.5, 2.7) {Existing: ``James is Engineer''};
\node[font=\small\sffamily\itshape, text=black!40] at (-4.5, 1.9) {(ignored)};
\node[cmpbox, fill=llmyellow!25] (llm1) at (-4.5, 0.8) {LLM (no context)};
\node[font=\small\sffamily\bfseries, text=fusionred!80] at (-4.5, -0.1) {$\downarrow$ ADD operation};
\node[badresult] (bad) at (-4.5, -1.8) {
\textcolor{fusionred}{\ding{55}} ``James is Sr. Engineer''\\
\textcolor{fusionred}{\ding{55}} ``James is Engineer''\\
\textit{\small Two conflicting memories!}
};

\draw[proarrow] (msg1.south) -- (llm1.north);
\draw[proarrow] (llm1.south) -- (bad.north);

\node[title] at (4.5, 5.5) {With Context Retrieval (Cognis)};

\node[cmpbox, fill=inputgray!15] (msg2) at (4.5, 4.3) {New: ``Promoted to Sr. Engineer''};
\node[cmpbox, fill=vdbpurple!20] (exist2) at (4.5, 2.7) {Retrieved: ``James is Engineer'' (id=12)};
\node[cmpbox, fill=llmyellow!25] (llm2) at (4.5, 0.8) {LLM (with context)};
\node[font=\small\sffamily\bfseries, text=processgreen!80] at (4.5, -0.1) {$\downarrow$ UPDATE replaces\_id=12};
\node[goodresult] (good) at (4.5, -1.8) {
\textcolor{processgreen}{\ding{51}} ``James is Sr. Engineer''\\
\hspace{1em}(replaces \#12)\\
\textit{\small Single consistent memory + history}
};

\draw[proarrow] (msg2.south) -- (exist2.north);
\draw[proarrow] (exist2.south) -- (llm2.north);
\draw[proarrow] (llm2.south) -- (good.north);

\draw[dashed, black!40, line width=1.5pt] (0, 6.0) -- (0, -3.0);

\end{tikzpicture}%
}
\caption{Context-aware extraction comparison. \textbf{Left}: Without context, the LLM creates conflicting memories (existing memory ignored). \textbf{Right}: With context retrieval, the LLM issues UPDATE operations maintaining consistency.}
\label{fig:context-aware}
\end{figure}
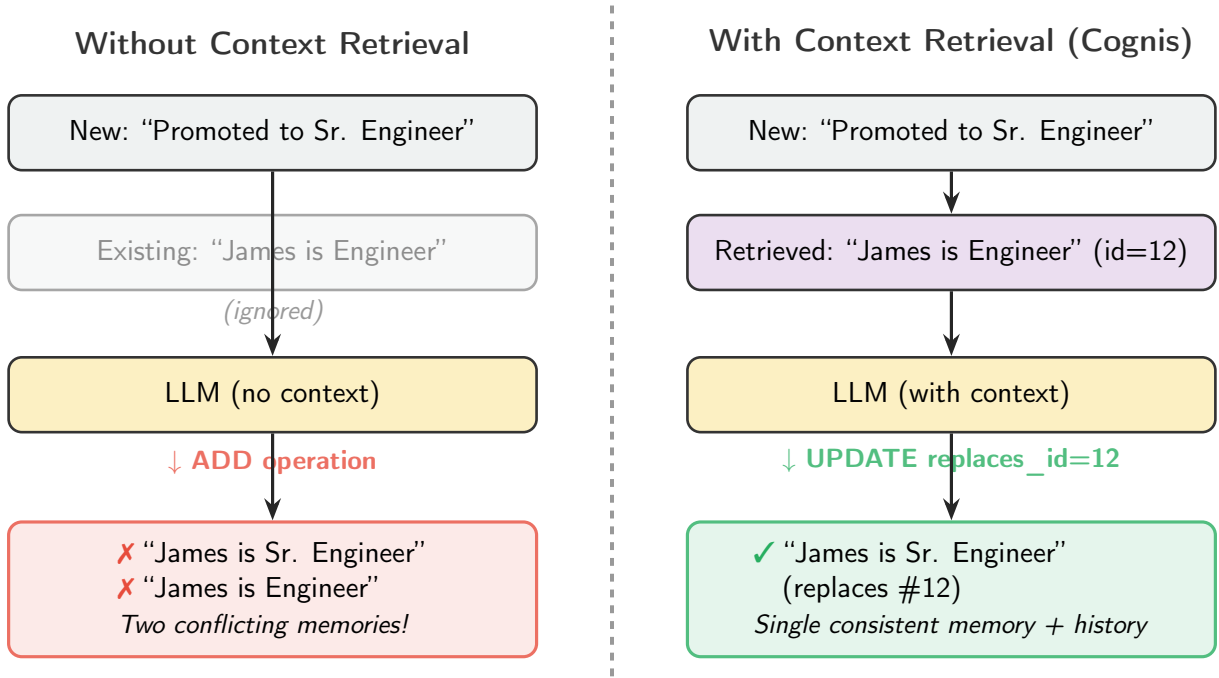

The core innovation of our ingestion pipeline is that fact extraction is \textit{not} performed in isolation. Before the LLM processes new messages, the system generates an embedding for the new message content, searches the VDB for the top-10 most similar existing memories, and provides both the new messages and existing memories to the LLM. Figure~\ref{fig:context-aware} illustrates why this context-aware approach matters. Without retrieving existing memories, the LLM treats each extraction independently, leading to conflicting or duplicate entries. With context retrieval, the system maintains a consistent, evolving knowledge base.

This context enables the LLM to make intelligent decisions about how to handle new information relative to what already exists in the memory store. Table~\ref{tab:operations} summarizes the four possible operations: \texttt{ADD} for genuinely new facts, \texttt{UPDATE} for information that supersedes existing memories (with version linking), \texttt{DELETE} for explicitly contradicted information, and \texttt{NONE} for duplicates or trivial content not worth storing.

\begin{table}[H]
\centering
\small
\caption{LLM Operation Decision Logic}
\label{tab:operations}
\begin{tabular}{lp{6.5cm}}
\toprule
\textbf{Operation} & \textbf{When Applied} \\
\midrule
\texttt{ADD} & New information not present in existing memories \\
\texttt{UPDATE} & New information supersedes or refines an existing memory (includes \texttt{replaces\_id} to link versions) \\
\texttt{DELETE} & New information explicitly contradicts or invalidates an existing memory \\
\texttt{NONE} & Information is a duplicate or not worth storing (e.g., small talk) \\
\bottomrule
\end{tabular}
\end{table}

For example, when processing the message ``I got promoted to Senior Engineer,'' the LLM returns structured JSON with the appropriate operations:

\begin{lstlisting}
{
  "operations": [
    {
      "action": "UPDATE",
      "fact": "James works at Google as a Senior Engineer",
      "replaces_id": 42,
      "category": "professional",
      "event_date": "2024-05-08"
    },
    {
      "action": "ADD",
      "fact": "James is excited about his new role",
      "category": "emotional"
    }
  ]
}
\end{lstlisting}

This approach prevents duplicate memories, maintains consistency when information changes, and ensures the memory store reflects the most current understanding of the user's world.

\subsection{Matryoshka Embedding Generation}

For extracted memories, we generate dual-dimension embeddings using the Matryoshka approach~\citep{matryoshka}. The system first generates a full 768D embedding with the prefix \texttt{search\_document:}, then truncates to 256D ($\mathbf{e}_{256} = \mathbf{e}_{768}[0:256]$), and finally re-normalizes the result ($\mathbf{e}_{256} = \frac{\mathbf{e}_{256}}{\|\mathbf{e}_{256}\|_2}$). This approach, pioneered by Matryoshka Representation Learning, ensures that the truncated embedding preserves semantic meaning while enabling faster approximate search. The 256D embedding captures the coarse-grained semantics sufficient for shortlisting, while the full 768D embedding provides fine-grained discrimination for accurate ranking.

\subsection{Dual-Store Persistence}

Each extracted memory is stored across both backends. \textbf{OpenSearch} stores a document containing full text content and metadata (category, scope), version tracking fields (\texttt{is\_current}, \texttt{replaces\_id}, \texttt{version}), temporal information (\texttt{event\_time}, \texttt{created\_at}), and BM25-indexed text fields with configurable analyzers. \textbf{VDB} stores two vector points with payloads: the 768D collection holds full embeddings for accurate retrieval, while the 256D collection contains truncated embeddings for fast shortlisting. Both collections include \texttt{memory\_id}, \texttt{user\_id}, \texttt{is\_current}, and \texttt{event\_time} in their payloads.

\subsection{Version Tracking and History}

Lyzr Cognis implements git-like automatic versioning with zero overhead. Every \texttt{UPDATE} operation automatically creates a new version while preserving full history---enabling time-travel capabilities for debugging and audit trails.

When the LLM returns an \texttt{UPDATE} operation, the system preserves history through version chaining. First, it marks the old memory by setting \texttt{is\_current=false} and \texttt{status=``historical''} in both OpenSearch and VDB. Then it creates a new memory with \texttt{replaces\_id} pointing to the old memory's ID, and increments the version number (\texttt{version = old\_version + 1}).

\begin{figure}[h]
\centering
\resizebox{\textwidth}{!}{%
\begin{tikzpicture}[
    versionbox/.style={rectangle, draw=black!80, rounded corners=6pt, minimum width=6.5cm, minimum height=2.2cm, align=left, font=\small\sffamily, line width=1pt, inner sep=10pt},
    currentbox/.style={rectangle, draw=processgreen!80, rounded corners=6pt, minimum width=6.5cm, minimum height=2.2cm, align=left, font=\small\sffamily, line width=2pt, inner sep=10pt},
    chainlabel/.style={font=\small\sffamily\bfseries, text=black!70},
    fieldlabel/.style={font=\footnotesize\ttfamily, text=black!80},
    statusbadge/.style={rectangle, draw=none, rounded corners=3pt, font=\scriptsize\sffamily\bfseries, inner sep=3pt},
    timeline/.style={->, >=Stealth, line width=1.5pt, black!50},
    chainlink/.style={->, >=Stealth, line width=2.5pt, vdbpurple!80, dashed}
]

\draw[timeline] (-1, -2.5) -- (22, -2.5);
\node[font=\small\sffamily\itshape, text=black!50] at (22.5, -2.5) {time};

\node[versionbox, fill=inputgray!15] (v1) at (2.5, 0) {
    \textbf{``James works as Software Engineer''}\\[4pt]
    \fieldlabel{id: 101}\\
    \fieldlabel{version: 1}\\
    \fieldlabel{replaces\_id: null}\\
    \fieldlabel{is\_current: false}
};
\node[statusbadge, fill=inputgray!40, text=black!70] at (2.5, 1.6) {HISTORICAL};
\node[chainlabel] at (2.5, 2.2) {v1};
\node[font=\scriptsize\sffamily, text=black!50] at (2.5, -2.0) {Jan 2023};

\node[versionbox, fill=inputgray!15] (v2) at (10, 0) {
    \textbf{``James works as Senior Engineer''}\\[4pt]
    \fieldlabel{id: 142}\\
    \fieldlabel{version: 2}\\
    \fieldlabel{replaces\_id: 101}\\
    \fieldlabel{is\_current: false}
};
\node[statusbadge, fill=inputgray!40, text=black!70] at (10, 1.6) {HISTORICAL};
\node[chainlabel] at (10, 2.2) {v2};
\node[font=\scriptsize\sffamily, text=black!50] at (10, -2.0) {Aug 2023};

\node[currentbox, fill=processgreen!12] (v3) at (17.5, 0) {
    \textbf{``James works as Tech Lead''}\\[4pt]
    \fieldlabel{id: 187}\\
    \fieldlabel{version: 3}\\
    \fieldlabel{replaces\_id: 142}\\
    \fieldlabel{is\_current: true}
};
\node[statusbadge, fill=processgreen!40, text=processgreen!90] at (17.5, 1.6) {CURRENT};
\node[chainlabel] at (17.5, 2.2) {v3};
\node[font=\scriptsize\sffamily, text=black!50] at (17.5, -2.0) {Feb 2024};

\draw[chainlink] (v2.west) -- (v1.east) node[midway, above, font=\small\sffamily\bfseries, text=vdbpurple!80] {replaces\_id};
\draw[chainlink] (v3.west) -- (v2.east) node[midway, above, font=\small\sffamily\bfseries, text=vdbpurple!80] {replaces\_id};

\node[rectangle, draw=apiblue!80, fill=apiblue!10, rounded corners=5pt, minimum width=8cm, inner sep=8pt, font=\small\sffamily, line width=1pt] (querybox) at (10, -4.2) {
    \textbf{Query:} ``What were all my previous jobs?'' $\rightarrow$ Returns: v1, v2, v3 (full history)
};

\end{tikzpicture}%
}
\caption{Version chaining for memory history. Each UPDATE creates a new version linked via \texttt{replaces\_id}. Historical versions have \texttt{is\_current=false}, enabling time-travel queries that traverse the chain to retrieve complete evolution of facts.}
\label{fig:version-chain}
\end{figure}
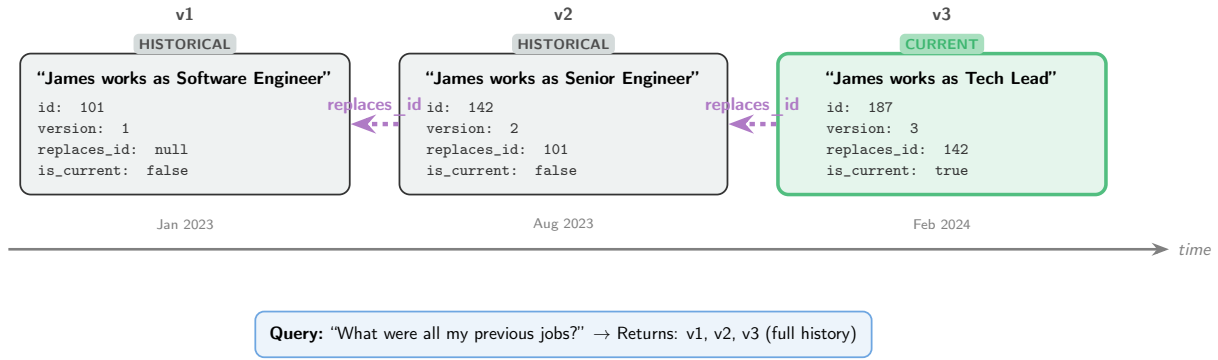

This approach provides several benefits: automatic versioning where updates or deletes create new versions without explicit user action; a full audit trail showing how information evolved over time; time-travel queries that retrieve any previous state (e.g., ``What were all my previous jobs?''); and rollback capability to debug what changed and when by traversing version chains.

This chain enables powerful historical queries. When a user asks ``What were all my previous jobs?'', the retrieval pipeline can traverse the version chain to return the complete employment history, not just the current position.

\section{Retrieval Pipeline}
\label{sec:retrieval}

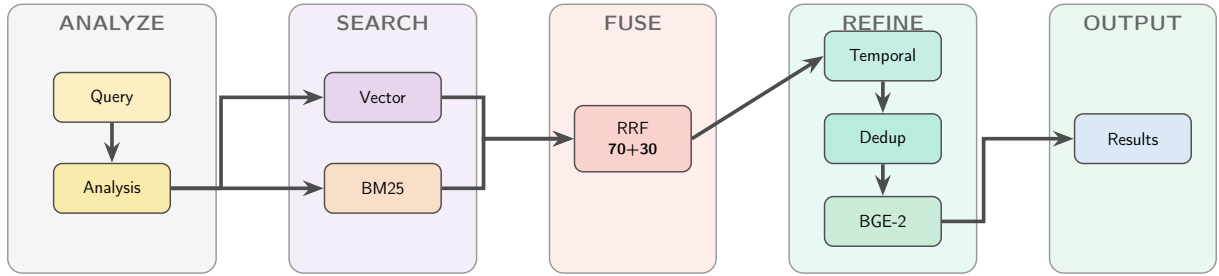
\begin{figure}[h]
\centering
\resizebox{\textwidth}{!}{%
\begin{tikzpicture}[
    node distance=1.0cm and 1.5cm,
    retbox/.style={rectangle, draw=black!70, rounded corners=6pt, minimum width=2.8cm, minimum height=1.2cm, align=center, font=\large\sffamily, line width=1pt},
    sectionbox/.style={rectangle, draw=black!40, rounded corners=10pt, inner sep=15pt, line width=1pt},
    bigarrow/.style={->, >=Stealth, line width=2.5pt, black!70}
]

\node[sectionbox, fill=inputgray!10, minimum width=5.0cm, minimum height=6.5cm] (analyzebox) at (2.5, 0) {};
\node[phaselabel, font=\Large\bfseries\sffamily, text=black!60] at (2.5, 2.8) {ANALYZE};

\node[sectionbox, fill=vdbpurple!10, minimum width=4.5cm, minimum height=6.5cm] (searchbox) at (9.0, 0) {};
\node[phaselabel, font=\Large\bfseries\sffamily, text=black!60] at (9.0, 2.8) {SEARCH};

\node[sectionbox, fill=fusionred!10, minimum width=4.0cm, minimum height=6.5cm] (fusebox) at (15.0, 0) {};
\node[phaselabel, font=\Large\bfseries\sffamily, text=black!60] at (15.0, 2.8) {FUSE};

\node[sectionbox, fill=boostcyan!10, minimum width=4.5cm, minimum height=6.5cm] (refinebox) at (21.0, 0) {};
\node[phaselabel, font=\Large\bfseries\sffamily, text=black!60] at (21.0, 2.8) {REFINE};

\node[sectionbox, fill=processgreen!10, minimum width=4.0cm, minimum height=6.5cm] (outputbox) at (27.0, 0) {};
\node[phaselabel, font=\Large\bfseries\sffamily, text=black!60] at (27.0, 2.8) {OUTPUT};

\node[retbox, fill=llmyellow!25] (query) at (2.5, 1.0) {Query};
\node[retbox, fill=llmyellow!35] (analyze) at (2.5, -1.2) {Analysis};

\node[retbox, fill=vdbpurple!25] (vector) at (9.0, 1.0) {Vector};
\node[retbox, fill=mongoorange!25] (bm25) at (9.0, -1.2) {BM25};

\node[retbox, fill=fusionred!25, minimum height=1.6cm] (rrf) at (15.0, 0) {RRF\\{\normalsize\bfseries 70+30}};

\node[retbox, fill=boostcyan!25] (temporal) at (21.0, 2.0) {Temporal};
\node[retbox, fill=boostcyan!30] (dedup) at (21.0, 0) {Dedup};
\node[retbox, fill=processgreen!25] (rerank) at (21.0, -2.0) {BGE-2};

\node[retbox, fill=apiblue!20] (output) at (27.0, 0) {Results};

\draw[bigarrow] (query) -- (analyze);

\draw[bigarrow] (analyze.east) -- ++(1.2, 0) |- (vector.west);
\draw[bigarrow] (analyze.east) -- ++(1.2, 0) |- (bm25.west);

\draw[bigarrow] (vector.east) -- ++(1.0, 0) |- (rrf.west);
\draw[bigarrow] (bm25.east) -- ++(1.0, 0) |- (rrf.west);

\draw[bigarrow] (rrf.east) -- (temporal.west);

\draw[bigarrow] (temporal) -- (dedup);
\draw[bigarrow] (dedup) -- (rerank);

\draw[bigarrow] (rerank.east) -- ++(1.0, 0) |- (output.west);

\end{tikzpicture}%
}
\caption{Retrieval pipeline: Query analysis $\rightarrow$ parallel Vector/BM25 search $\rightarrow$ RRF fusion (70\%+30\%) $\rightarrow$ temporal boost, dedup $\rightarrow$ BGE-2 rerank $\rightarrow$ results.}
\label{fig:retrieval}
\end{figure}

The retrieval pipeline implements a sophisticated hybrid search combining vector similarity and keyword matching, with temporal boosting and a BGE-2 cross-encoder reranker for final refinement (Figure~\ref{fig:retrieval}).

\subsection{Query Analysis}

Before search execution, we analyze the query to determine retrieval strategy:

\textbf{Temporal Intent Detection}: Keywords like ``when'', ``yesterday'', ``last week'', ``on May 8th'' trigger temporal boosting. The system extracts the time reference and calculates appropriate time windows for relevance scoring.

\textbf{History Detection}: Keywords like ``previous'', ``all my'', ``history'', ``journey'', ``over time'' indicate historical queries that should include superseded memories (\texttt{is\_current=false}), enabling retrieval of complete version chains.

This analysis shapes both the search filters and post-processing stages.

\subsection{Matryoshka Two-Stage Vector Search}

We implement efficient two-stage retrieval using Matryoshka embeddings, trading off speed and accuracy. \textbf{Stage 1 (Fast Shortlisting)} searches the 256D collection for 200 candidate memories with $\sim$5-10ms latency, using coarse-grained semantics for rapid filtering. \textbf{Stage 2 (Accurate Re-ranking)} filters the 768D collection by the shortlist memory IDs and computes precise similarity with full embeddings, achieving high-precision semantic discrimination in $\sim$10-20ms.

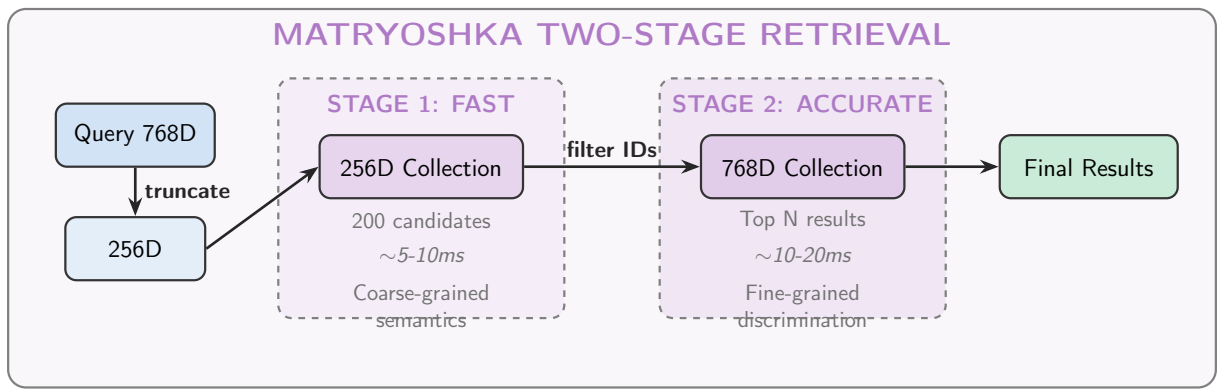
\begin{figure}[h]
\centering
\resizebox{\textwidth}{!}{%
\begin{tikzpicture}[
    node distance=1.0cm and 1.5cm,
    matbox/.style={rectangle, draw=black!80, rounded corners=5pt, minimum width=3.0cm, minimum height=1.0cm, align=center, line width=1pt, font=\normalsize\sffamily}
]

\node[procontainer, fill=vdbpurple!5, minimum width=19cm, minimum height=6.0cm] (mainbox) at (9, 0) {};
\node[font=\Large\bfseries\sffamily, text=vdbpurple!80] at (9, 2.6) {MATRYOSHKA TWO-STAGE RETRIEVAL};

\node[matbox, fill=apiblue!25, minimum width=2.5cm] (query) at (1.5, 1.0) {Query 768D};
\node[matbox, fill=apiblue!15, minimum width=2.2cm] (trunc) at (1.5, -0.8) {256D};
\draw[proarrow] (query) -- node[right, font=\small\sffamily\bfseries] {truncate} (trunc);

\node[stagebox, fill=vdbpurple!10, minimum width=4.5cm, minimum height=3.8cm, inner sep=10pt] (s1box) at (6.0, 0) {};
\node[phaselabel, text=vdbpurple!80, font=\normalsize\bfseries\sffamily] at (6.0, 1.5) {STAGE 1: FAST};
\node[matbox, fill=vdbpurple!25, minimum width=3.2cm] (stage1) at (6.0, 0.5) {256D Collection};
\node[infolabel, font=\small\sffamily] at (6.0, -0.35) {200 candidates};
\node[infolabel, font=\small\sffamily\itshape] at (6.0, -0.9) {$\sim$5-10ms};

\node[stagebox, fill=vdbpurple!15, minimum width=4.5cm, minimum height=3.8cm, inner sep=10pt] (s2box) at (12.0, 0) {};
\node[phaselabel, text=vdbpurple!80, font=\normalsize\bfseries\sffamily] at (12.0, 1.5) {STAGE 2: ACCURATE};
\node[matbox, fill=vdbpurple!30, minimum width=3.2cm] (stage2) at (12.0, 0.5) {768D Collection};
\node[infolabel, font=\small\sffamily] at (12.0, -0.35) {Top N results};
\node[infolabel, font=\small\sffamily\itshape] at (12.0, -0.9) {$\sim$10-20ms};

\node[matbox, fill=processgreen!25, minimum width=2.8cm] (output) at (16.5, 0.5) {Final Results};

\draw[proarrow] (trunc.east) -- (stage1.west);
\draw[proarrow] (stage1.east) -- (stage2.west) node[midway, above, font=\small\sffamily\bfseries] {filter IDs};
\draw[proarrow] (stage2.east) -- (output.west);

\node[infolabel, align=center, font=\small\sffamily] at (6.0, -1.7) {Coarse-grained\\semantics};
\node[infolabel, align=center, font=\small\sffamily] at (12.0, -1.7) {Fine-grained\\discrimination};

\end{tikzpicture}%
}
\caption{Matryoshka two-stage retrieval: truncated 256D embeddings enable fast shortlisting (200 candidates in $\sim$5ms), followed by accurate 768D re-ranking ($\sim$15ms) for final results.}
\label{fig:matryoshka}
\end{figure}

Two-stage Matryoshka retrieval provides substantial latency improvements with minimal accuracy impact. In our benchmarks, p50 latency drops from 32ms (single-stage 768D) to 18ms (256D shortlist + 768D filter), a 44\% reduction. The p99 latency improves from 89ms to 51ms, a 43\% reduction that matters for tail latency in production systems. The accuracy cost is minimal: in our testing, temporal F1 drops by only 1.4\% when using single-stage 768D search, confirming that the 256D coarse filtering preserves the candidates that matter for final ranking.

\subsection{BM25 Text Search}

OpenSearch provides native BM25 search that complements semantic similarity. BM25 excels at exact name matching (``James'', ``Google''), technical terms and acronyms, dates and numbers, and rare words with high discriminative power. Our ablation studies (Section~\ref{sec:ablation}) demonstrate that OpenSearch's native BM25 implementation with configurable text analyzers significantly outperforms MongoDB's text indexing, yielding a +20.3\% improvement on open-domain LLM Judge scores. The query is executed against OpenSearch's BM25 index with user isolation:

\begin{lstlisting}
{
  "query": {
    "bool": {
      "must": {"match": {"content": query}},
      "filter": [
        {"term": {"owner_id": user_id}},
        {"term": {"is_current": true}}
      ]
    }
  }
}
\end{lstlisting}

\subsection{Reciprocal Rank Fusion}

We combine results from vector and BM25 search using Reciprocal Rank Fusion (RRF)~\citep{cormack2009rrf}:

\begin{equation}
\text{RRF}(d) = \sum_{r \in R} \frac{1}{k + \text{rank}_r(d)}
\end{equation}

where $k = 10$ is a smoothing constant and $\text{rank}_r(d)$ is the rank of document $d$ in retriever $r$.

The final fused score combines weighted contributions:

\begin{equation}
\text{score}_{\text{fused}} = 0.70 \cdot \text{RRF}_{\text{vector}} + 0.30 \cdot \text{RRF}_{\text{BM25}}
\end{equation}

The 70/30 vector-BM25 weighting reflects an empirical observation about query types in conversational memory. The majority of queries seek semantic similarity (``Tell me about my hobbies'', ``What do I like to do?'') where vector search excels at matching paraphrased concepts. However, approximately 30\% of queries contain specific anchors---names (``What did Sarah say?''), dates (``meeting on March 15th''), technical terms (``my AWS credentials'')---where BM25's exact matching provides critical signal that embeddings may dilute.

We evaluated alternative weightings during development: equal weighting (50/50) over-weighted keyword matches for paraphrase queries, causing irrelevant memories with incidental term overlap to outrank semantically relevant ones. Conversely, 80/20 weighting missed important exact matches when users queried specific entities by name. The 70/30 balance consistently outperformed alternatives across all four question types in our validation set, achieving the best trade-off between semantic coverage and lexical precision.

\subsection{Temporal Boosting}

For queries with temporal intent, we apply time-based relevance scoring. The temporal score measures how close a memory's event time is to the query's temporal reference:

\begin{equation}
\text{temporal\_score} = \max\left(0.1, 1 - \frac{|\text{event\_time} - \text{query\_date}|}{\text{window\_days}}\right)
\end{equation}

The final score integrates temporal relevance:

\begin{equation}
\text{score}_{\text{final}} = 0.60 \cdot \text{score}_{\text{fused}} + 0.40 \cdot \text{temporal\_score}
\end{equation}

This boosting is crucial for questions like ``What did I do last Tuesday?'' where temporal proximity should outweigh pure semantic similarity.

\subsection{Content Deduplication}

Post-fusion, we remove near-duplicate memories to ensure diversity in results. Two memories are considered duplicates if their semantic similarity score exceeds 99\%. When duplicates are found, only the highest-scoring version is retained.

This is particularly important when version chains exist---without deduplication, both current and historical versions of the same fact might appear in results.

\subsection{BGE-2 Cross-Encoder Reranking}

As a final refinement stage, we apply a cross-encoder reranker to the top candidates. The BGE-2 reranker~\citep{bge2024} jointly encodes the query and each candidate document, providing more accurate relevance scores than bi-encoder similarity alone.

The reranker operates via a remote HTTP endpoint, sending the query and top-N candidates (post-deduplication) to the reranker service, receiving refined relevance scores for each candidate, and re-sorting results by these scores. This stage adds approximately 20-50ms latency but provides significant quality improvements, particularly for nuanced queries where initial retrieval may not perfectly order results.

\subsection{Historical Query Handling}

When \texttt{include\_historical=True} (detected via query analysis), the pipeline adjusts its behavior by removing the \texttt{is\_current=true} filter from both vector and BM25 searches, including both current and superseded memories in results, traversing version chains via \texttt{replaces\_id} links, and sorting results chronologically (oldest first) to show evolution. This enables queries like ``What were all my previous jobs?'' to return the complete employment history with version relationships.

\section{Experimental Setup}
\label{sec:setup}

\subsection{Datasets}

We evaluate Cognis on two complementary benchmarks that test different aspects of long-term memory.

\subsubsection{LoCoMo}

The \textbf{LoCoMo} (Long-Context Modeling) benchmark~\citep{locomo} tests memory systems across multi-session conversations requiring recall across 50+ sessions. LoCoMo provides four distinct question categories that test different memory capabilities: \textbf{Single-Hop} questions require direct fact recall from a single session (e.g., ``What is James's favorite sport?''); \textbf{Multi-Hop} questions involve reasoning across multiple facts (e.g., ``What indoor activity would Alice enjoy with her dog?''); \textbf{Open Domain} questions are broad and require comprehensive recall (e.g., ``Tell me about James's hobbies''); and \textbf{Temporal} questions are time-sensitive (e.g., ``What did James do last Tuesday?''). This categorization enables fine-grained analysis of where memory systems succeed or struggle.

\subsubsection{LongMemEval}

The \textbf{LongMemEval} benchmark~\citep{longmemeval} is a comprehensive test of chat assistant long-term memory comprising 500 questions spanning 6 question types. LongMemEval defines six question types that systematically test different memory capabilities. \textbf{SS-User} questions test recall of single-session user-stated facts (e.g., ``What city did I say I moved to?''). \textbf{SS-Assistant} questions test recall of assistant responses (e.g., ``What recipe did you recommend last time?''). \textbf{SS-Preference} questions test user preference recall (e.g., ``Do you remember my dietary restrictions?''). \textbf{Multi-Session} questions require cross-conversation reasoning, connecting information scattered across separate interactions. \textbf{Temporal Reasoning} questions test time-dependent queries where the answer depends on when events occurred relative to each other. \textbf{Knowledge Update} questions test handling of changed information, where a user corrects or updates a previously stated fact.

The key methodological difference from LoCoMo is that LongMemEval's multi-session design tests whether memory systems can maintain coherence across conversation boundaries---a challenge that exercises both ingestion (correctly associating facts across sessions) and retrieval (bridging semantic gaps between separate conversations) capabilities.

\subsection{Evaluation Methodology}

Both benchmarks share a 5-phase evaluation pipeline: \textbf{INGEST} loads conversation sessions into each memory system; \textbf{INDEXING} waits for providers to complete extraction and embedding; \textbf{SEARCH} queries memories for each evaluation question; \textbf{ANSWER} generates answers using retrieved memories; and \textbf{EVALUATE} scores answers using question type-specific LLM judges. We use specialized judge prompts for different question types (see Appendix~\ref{app:prompts}), including off-by-one tolerance for temporal questions, rubric-based scoring for preference questions, and appropriate handling of knowledge updates. LoCoMo uses GPT-4 for answer generation; LongMemEval uses GPT-4.1 for answer generation to isolate memory architecture differences from LLM capability variation.

\subsection{Metrics}

\subsubsection{LoCoMo Metrics}

We report three complementary metrics that capture different aspects of answer quality. \textbf{F1 Score} measures token-level precision and recall between generated and ground truth answers, providing a strict assessment of factual content overlap. \textbf{BLEU-1 (B1)} captures unigram lexical similarity, offering a softer measure of word-level correspondence. Finally, \textbf{LLM Judge (J)} uses GPT-4 to evaluate answer correctness on a 0-100 scale, capturing semantic equivalence beyond surface-level lexical matching---particularly important when correct answers may be paraphrased or expressed differently than the reference.

\subsubsection{LongMemEval Metrics}

We report accuracy as the primary metric, measuring end-to-end correctness. Retrieval quality is independently assessed via LLM-based chunk relevance evaluation, supplemented by Hit@K, Mean Reciprocal Rank (MRR), Normalized Discounted Cumulative Gain (NDCG), Precision@K, Recall@K, and F1@K.

\subsection{Baselines}

\subsubsection{LoCoMo Baselines}

We compare against 11 systems spanning different approaches to LLM memory. These include the \textbf{LoCoMo}~\citep{locomo} benchmark's baseline retrieval approach; document-focused methods like \textbf{ReadAgent}~\citep{readagent} with its gist-based memory and \textbf{MemoryBank}~\citep{memorybank} with memory consolidation; architecture-driven approaches including \textbf{MemGPT}~\citep{memgpt} (OS-inspired memory paging), \textbf{A-Mem}~\citep{amem} (agentic memory with structured management), and \textbf{A-Mem*} (A-Mem variant with LLM-as-a-Judge evaluation); framework integrations such as \textbf{LangMem} (LangChain-based) and \textbf{OpenAI}'s memory API; dedicated memory services \textbf{Zep}~\citep{zep} (long-term memory with session management) and \textbf{Mem0}~\citep{mem0} (vector-based with automatic extraction); and \textbf{Mem0g}, an enhanced variant of Mem0 incorporating graph-based knowledge representation for improved multi-hop reasoning.

\subsubsection{LongMemEval Baselines}

We compare against two competitive systems: \textbf{SuperMemory}~\citep{supermemory}, which focuses on knowledge graph integration for multi-hop reasoning, and \textbf{Zep/Graphiti}~\citep{zep}, which combines long-term memory with graph-based knowledge representation. All systems use GPT-4.1 for answer generation.

\section{Results on LoCoMo}
\label{sec:locomo-results}

\subsection{Main Results}

Table~\ref{tab:main-results} presents comprehensive results on the LoCoMo benchmark across all four question types. We report F1 score (F1), BLEU-1 (B1), and LLM-as-a-Judge score (J) where available, with higher values indicating better performance ($\uparrow$).

\begin{table*}[htbp]
\centering
\caption{Performance comparison on LoCoMo benchmark across question types. Metrics: F1 score, BLEU-1 (B1), LLM-as-a-Judge (J). Best in \textbf{bold}, second-best \underline{underlined}. ($\uparrow$) = higher is better.}
\label{tab:main-results}
\resizebox{\textwidth}{!}{%
\begin{tabular}{l|ccc|ccc|ccc|ccc}
\toprule
& \multicolumn{3}{c|}{\textbf{Single-Hop}} & \multicolumn{3}{c|}{\textbf{Multi-Hop}} & \multicolumn{3}{c|}{\textbf{Open-Domain}} & \multicolumn{3}{c}{\textbf{Temporal}} \\
\textbf{Method} & F1$\uparrow$ & B1$\uparrow$ & J$\uparrow$ & F1$\uparrow$ & B1$\uparrow$ & J$\uparrow$ & F1$\uparrow$ & B1$\uparrow$ & J$\uparrow$ & F1$\uparrow$ & B1$\uparrow$ & J$\uparrow$ \\
\midrule
LoCoMo & 25.02 & 19.75 & -- & 12.04 & 11.16 & -- & 40.36 & 29.05 & -- & 18.41 & 14.77 & -- \\
ReadAgent & 9.15 & 6.48 & -- & 5.31 & 5.12 & -- & 9.67 & 7.66 & -- & 12.60 & 8.87 & -- \\
MemoryBank & 5.00 & 4.77 & -- & 5.56 & 5.94 & -- & 6.61 & 5.16 & -- & 9.68 & 6.99 & -- \\
MemGPT & 26.65 & 17.72 & -- & 9.15 & 7.44 & -- & 41.04 & 34.34 & -- & 25.52 & 19.44 & -- \\
A-Mem & 27.02 & 20.09 & -- & 12.14 & 12.00 & -- & 44.65 & 37.06 & -- & 45.85 & 36.67 & -- \\
A-Mem* & 20.76 & 14.90 & 39.79 & 9.22 & 8.81 & 18.85 & 33.34 & 27.58 & 54.05 & 35.40 & 31.08 & 49.91 \\
LangMem & 35.51 & 26.86 & 62.23 & 26.04 & \underline{22.32} & 47.92 & 40.91 & 33.63 & 71.12 & 30.75 & 25.84 & 23.43 \\
Zep & 35.74 & 23.30 & 61.70 & 19.37 & 14.82 & 41.35 & \underline{49.56} & 38.92 & \underline{76.60} & 42.00 & 34.53 & 49.31 \\
OpenAI & 34.30 & 23.72 & 63.79 & 20.09 & 15.42 & 42.92 & 39.31 & 31.16 & 62.29 & 14.04 & 11.25 & 21.71 \\
Mem0 & \underline{38.72} & \underline{27.13} & \underline{67.13} & \underline{28.64} & 21.58 & \textbf{51.15} & 47.65 & 38.72 & 72.93 & 48.93 & \underline{40.51} & 55.51 \\
Mem0g & 38.09 & 26.03 & 65.71 & 24.32 & 18.82 & 47.19 & 49.27 & \underline{40.30} & 75.71 & \underline{51.55} & 40.28 & \underline{58.13} \\
\midrule
\textbf{Lyzr Cognis} & \textbf{48.66} & \textbf{35.76} & \textbf{71.99} & \textbf{31.51} & \textbf{23.22} & 51.04 & \textbf{54.77} & \textbf{45.12} & \textbf{85.85} & \textbf{62.68} & \textbf{58.95} & \textbf{77.26} \\
\midrule
\multicolumn{1}{r|}{$\Delta$ vs best} & \textcolor{green!60!black}{+25.7\%} & & \textcolor{green!60!black}{+7.2\%} & \textcolor{green!60!black}{+10.0\%} & & -- & \textcolor{green!60!black}{+10.5\%} & & \textcolor{green!60!black}{+12.1\%} & \textcolor{green!60!black}{+21.6\%} & & \textcolor{green!60!black}{+32.9\%} \\
\bottomrule
\end{tabular}%
}
\end{table*}

\textbf{Single-hop questions.} Cognis achieves 48.66 F1 (+25.7\% over Mem0), demonstrating that context-aware extraction prevents memory pollution and that OpenSearch's native BM25 significantly improves entity-specific recall. The LLM Judge improvement (+7.2\%) confirms semantic retrieval quality beyond lexical matching.

\textbf{Multi-hop questions.} Cognis achieves 31.51 F1 (+10.0\% over Mem0). Our hybrid retrieval successfully surfaces related facts by combining semantic similarity with OpenSearch BM25 keyword matching. The gain reflects improved cross-fact retrieval through better term matching, though multi-hop reasoning remains an inherently difficult open challenge.

\textbf{Open-domain questions.} Cognis achieves 54.77 F1 (+10.5\% over Zep) and a striking 85.85 LLM Judge (+12.1\% over Zep's 76.60). This represents a dramatic improvement: Cognis now leads on \textit{both} F1 and LLM Judge for open-domain questions. OpenSearch's native BM25 with configurable text analysis is the key driver---broad entity queries benefit substantially from proper tokenization and term-frequency weighting that MongoDB's text indexing could not provide.

\textbf{Temporal questions.} Our strongest F1 gains: 62.68 F1 (+21.6\% over Mem0g) and 77.26 LLM Judge (+32.9\%). This validates Cognis's temporal boosting mechanism, which adjusts scores based on proximity between query references and memory event times. The BLEU-1 score of 58.95 (+45.5\% over Mem0's 40.51) indicates substantially better lexical coverage in temporal answer generation.

\subsection{Error Analysis}
\label{sec:error-analysis}

To understand where Cognis still fails, we manually analyzed 50 incorrect temporal question responses---the category where we see the largest gains but also the most room for improvement.

\textbf{Error distribution}: Of the analyzed failures, 34\% stemmed from retrieval failures (correct memory exists but was not retrieved in top-K), 28\% from temporal boosting mismatch (wrong memory boosted due to date proximity), 22\% from reranker ordering errors (correct memory retrieved but ranked below incorrect alternatives), and 16\% from LLM reasoning errors (correct memories retrieved and ranked, but answer generation failed).

\textbf{Illustrative failure case}: Consider the query ``What did James eat for breakfast last Tuesday?'' The system retrieved ``James enjoys oatmeal for breakfast'' (a general preference stored months earlier) rather than ``James had eggs and toast on Tuesday morning'' (the specific event from the target date). The temporal boosting mechanism correctly identified the query as time-sensitive, but it boosted the preference fact because that memory document happened to be created closer to the query date---even though the query sought a specific dated event.

This reveals a limitation in Cognis's current temporal boosting: it operates on memory \textit{storage timestamps} rather than \textit{event timestamps}. When the event date is explicitly mentioned in the memory content (e.g., ``on Tuesday morning''), the system should use that date for temporal boosting rather than the storage date. Improving event date extraction during ingestion would prevent this class of errors.

\subsection{Ablation Studies}
\label{sec:ablation}

We conduct two sets of ablation experiments to quantify the contribution of individual architectural components: (1) the impact of embedding model choice on retrieval quality, and (2) the impact of retrieval pipeline variants including storage backend, reranker choice, and query decomposition strategy.

\subsubsection{Embedding Model Comparison}

Table~\ref{tab:embedding-ablation} compares four embedding models within our retrieval pipeline, holding all other components constant. Each model generates dual-dimension (768D + 256D) Matryoshka embeddings for two-stage retrieval.

\begin{table*}[htbp]
\centering
\caption{Embedding model ablation on LoCoMo benchmark. All configurations use the same retrieval pipeline (RRF fusion + BGE-2 reranking + OpenSearch BM25). Best per-column in \textbf{bold}.}
\label{tab:embedding-ablation}
\resizebox{\textwidth}{!}{%
\begin{tabular}{l|ccc|ccc|ccc|ccc}
\toprule
& \multicolumn{3}{c|}{\textbf{Single-Hop}} & \multicolumn{3}{c|}{\textbf{Multi-Hop}} & \multicolumn{3}{c|}{\textbf{Open-Domain}} & \multicolumn{3}{c}{\textbf{Temporal}} \\
\textbf{Embedding Model} & F1$\uparrow$ & B1$\uparrow$ & J$\uparrow$ & F1$\uparrow$ & B1$\uparrow$ & J$\uparrow$ & F1$\uparrow$ & B1$\uparrow$ & J$\uparrow$ & F1$\uparrow$ & B1$\uparrow$ & J$\uparrow$ \\
\midrule
Nomic Embed & \textbf{50.23} & \textbf{41.26} & 69.15 & 29.02 & 21.17 & 52.08 & 50.72 & 41.48 & 68.61 & \textbf{60.48} & \textbf{52.69} & 72.59 \\
Gemma Embed & 40.94 & 30.88 & \textbf{71.88} & 23.22 & 14.72 & \textbf{69.23} & 44.85 & 33.37 & 58.33 & 54.01 & 42.46 & 70.27 \\
Nomic V2 & 42.38 & 31.42 & 65.96 & \textbf{30.31} & \textbf{23.71} & 50.00 & \textbf{51.67} & \textbf{42.44} & 68.12 & 58.54 & 51.23 & 73.21 \\
Jina V3 & 41.42 & 30.10 & 65.25 & 30.41 & 22.81 & 52.08 & 48.66 & 40.19 & 67.18 & 59.37 & 51.85 & \textbf{74.77} \\
\bottomrule
\end{tabular}%
}
\end{table*}

\textbf{Key findings.} Embedding model choice has a pronounced and category-dependent impact on retrieval quality. \textit{Nomic Embed} achieves the highest single-hop F1 (50.23) and temporal F1 (60.48), excelling at direct fact recall and time-sensitive queries. \textit{Gemma Embed} shows a striking F1-Judge divergence: despite the lowest single-hop F1 (40.94), it achieves the highest Judge scores for both single-hop (71.88) and multi-hop (69.23), suggesting that its embeddings capture semantic correctness better than token-level overlap. \textit{Nomic V2} leads on multi-hop (F1=30.31) and open-domain (F1=51.67), demonstrating stronger cross-fact retrieval. \textit{Jina V3} delivers balanced performance without leading any single category.

These results suggest that embedding model selection creates a fundamental tradeoff between token-level precision (F1) and semantic correctness (Judge), and that task-specific embedding selection---or ensemble strategies---may yield further improvements.

\subsubsection{Retrieval Pipeline Ablation}

Table~\ref{tab:pipeline-ablation} isolates the impact of three architectural choices: BM25 storage backend (MongoDB vs.\ OpenSearch), cross-encoder reranker (BGE-2 vs.\ Zero Entropy), and query preprocessing (direct vs.\ LLM decomposition).

\begin{table*}[htbp]
\centering
\caption{Retrieval pipeline ablation on LoCoMo benchmark. All configurations use the same embedding model. Best per-column in \textbf{bold}.}
\label{tab:pipeline-ablation}
\resizebox{\textwidth}{!}{%
\begin{tabular}{l|ccc|ccc|ccc|ccc}
\toprule
& \multicolumn{3}{c|}{\textbf{Single-Hop}} & \multicolumn{3}{c|}{\textbf{Multi-Hop}} & \multicolumn{3}{c|}{\textbf{Open-Domain}} & \multicolumn{3}{c}{\textbf{Temporal}} \\
\textbf{Configuration} & F1$\uparrow$ & B1$\uparrow$ & J$\uparrow$ & F1$\uparrow$ & B1$\uparrow$ & J$\uparrow$ & F1$\uparrow$ & B1$\uparrow$ & J$\uparrow$ & F1$\uparrow$ & B1$\uparrow$ & J$\uparrow$ \\
\midrule
BGE-2 Reranker (MongoDB) & 45.33 & 33.96 & 70.21 & 30.55 & \textbf{23.69} & 51.04 & 53.76 & 44.65 & 71.34 & 60.86 & 53.43 & 76.64 \\
Zero Entropy Reranker & 45.95 & 34.86 & 68.23 & 27.50 & 20.75 & 46.97 & 52.11 & 43.98 & 69.59 & 59.73 & 58.76 & 76.17 \\
LLM Query Decomposition & 44.73 & 34.86 & 69.52 & 29.50 & 21.09 & 50.79 & 46.11 & 37.73 & 68.45 & 60.75 & 57.65 & \textbf{77.15} \\
\textbf{BGE-2 + OpenSearch} & \textbf{48.66} & \textbf{35.76} & \textbf{71.99} & \textbf{31.51} & 23.22 & \textbf{51.04} & \textbf{54.77} & \textbf{45.12} & \textbf{85.85} & \textbf{62.68} & \textbf{58.95} & 77.26 \\
\bottomrule
\end{tabular}%
}
\end{table*}

\textbf{OpenSearch vs.\ MongoDB BM25.} The single largest architectural improvement comes from switching the BM25 backend from MongoDB to OpenSearch. Comparing the BGE-2 (MongoDB) row with BGE-2 + OpenSearch, we observe: +7.3\% single-hop F1 (45.33$\rightarrow$48.66), +3.1\% multi-hop F1 (30.55$\rightarrow$31.51), and most dramatically, +20.3\% open-domain Judge (71.34$\rightarrow$85.85). OpenSearch's native BM25 implementation with configurable text analyzers provides substantially better term matching for broad entity queries, where proper tokenization and term-frequency weighting are critical.

\textbf{Reranker comparison.} BGE-2 outperforms the Zero Entropy reranker on multi-hop F1 (30.55 vs.\ 27.50, $\Delta$=+11.1\%) and single-hop Judge (70.21 vs.\ 68.23), while Zero Entropy shows slightly higher single-hop F1 (45.95 vs.\ 45.33) and temporal BLEU-1 (58.76 vs.\ 53.43). BGE-2's advantage on multi-hop reasoning---where nuanced relevance discrimination between related facts is critical---motivated its selection as the default reranker.

\textbf{LLM query decomposition.} Decomposing queries into sub-questions via an LLM before retrieval yields marginal gains on temporal Judge (77.15, the highest in this column) but consistently decreases performance on simpler question types: $-$1.3\% single-hop F1, $-$14.2\% open-domain F1. This suggests that query decomposition introduces retrieval noise for well-formed queries where the original query already captures intent precisely. The technique may be better suited as a selective strategy triggered by query complexity heuristics rather than a default pipeline stage.

\subsubsection{RRF Weight Ablation}

To validate the 70/30 vector/BM25 weighting used throughout our evaluation, we conducted a systematic ablation varying the RRF weight distribution across seven configurations (30/70 through 80/20). All other pipeline parameters---embedding model, reranker (BGE-2), temporal boosting, top\_k=50---were held constant. Table~\ref{tab:rrf-ablation} presents the full results.

\begin{table*}[htbp]
\centering
\caption{RRF weight ablation on LoCoMo benchmark. Seven vector/BM25 weight configurations evaluated with all other pipeline components held constant. Best per-column in \textbf{bold}.}
\label{tab:rrf-ablation}
\resizebox{\textwidth}{!}{%
\begin{tabular}{l|ccc|ccc|ccc|ccc}
\toprule
& \multicolumn{3}{c|}{\textbf{Single-Hop}} & \multicolumn{3}{c|}{\textbf{Multi-Hop}} & \multicolumn{3}{c|}{\textbf{Open-Domain}} & \multicolumn{3}{c}{\textbf{Temporal}} \\
\textbf{Config (Vec/BM25)} & F1$\uparrow$ & B1$\uparrow$ & J$\uparrow$ & F1$\uparrow$ & B1$\uparrow$ & J$\uparrow$ & F1$\uparrow$ & B1$\uparrow$ & J$\uparrow$ & F1$\uparrow$ & B1$\uparrow$ & J$\uparrow$ \\
\midrule
\textbf{70/30} & 48.66 & 35.76 & 71.99 & \textbf{31.51} & \textbf{23.22} & \textbf{51.04} & 54.77 & 45.12 & 85.85 & \textbf{62.68} & \textbf{58.95} & \textbf{77.26} \\
75/25 & \textbf{49.94} & \textbf{36.84} & \textbf{72.53} & 30.82 & 22.43 & 50.17 & 55.21 & 46.92 & 86.11 & 62.11 & 58.84 & 77.26 \\
80/20 & 49.82 & 36.41 & 72.08 & 29.87 & 22.04 & 49.62 & \textbf{55.34} & 46.08 & \textbf{86.74} & 61.21 & 57.33 & 74.94 \\
60/40 & 49.11 & 35.94 & 71.88 & 28.63 & 20.71 & 48.37 & 55.02 & 45.96 & 86.28 & 60.73 & 55.82 & 74.41 \\
50/50 & 49.44 & 36.08 & 70.11 & 27.18 & 19.44 & 45.72 & 52.18 & 40.63 & 83.02 & 60.58 & 55.91 & 74.08 \\
40/60 & 44.73 & 33.62 & 69.54 & 27.92 & 20.12 & 44.83 & 49.47 & 38.27 & 78.56 & 54.01 & 42.46 & 70.27 \\
30/70 & 44.03 & 34.88 & 69.67 & 27.66 & 19.08 & 45.91 & 48.63 & 38.51 & 76.69 & 52.45 & 40.79 & 69.47 \\
\bottomrule
\end{tabular}%
}
\end{table*}

\textbf{70/30 leads on the hardest categories.} The two most challenging question categories---multi-hop reasoning and temporal reasoning---are where RRF weight distribution has the most impact. The 70/30 configuration achieves the highest multi-hop F1 (31.51), the highest temporal F1 (62.68), and the highest temporal Judge score (77.26). These categories require the system to connect facts across multiple memories and reason about time-dependent information, precisely the scenarios where the balance between semantic linking (vector-dominant) and lexical anchoring (BM25 at 30\%) is critical. The 30\% BM25 weight provides sufficient keyword anchoring for entity names and temporal expressions while preserving the vector component's cross-memory semantic reasoning capacity.

\textbf{Overall variation is within noise, but per-category breakdown reveals 70/30's edge.} Across all seven configurations, single-hop F1 ranges from 44.03 to 49.94 ($\sim$6 point spread), while multi-hop F1 ranges from 27.18 to 31.51 ($\sim$4 point spread). The relatively flat overall performance confirms that the BGE-2 reranker stabilizes end-to-end quality regardless of initial weight distribution. However, the per-category breakdown reveals that 70/30 uniquely excels on the categories with the widest performance variance: multi-hop and temporal. The nearest competitor, 75/25, trails 70/30 on multi-hop F1 by 0.69 points and on temporal F1 by 0.57 points---small but consistent margins that compound across question types.

\textbf{Degradation at extremes.} BM25-heavy configurations (40/60, 30/70) show pronounced degradation: single-hop F1 drops by 8--10\% (from 48.66 to 44.03--44.73), and open-domain Judge scores fall by 7--9 points (from 85.85 to 76.69--78.56). This occurs because excessive BM25 weight overwhelms the semantic signal needed for paraphrase matching and broad entity queries. Conversely, vector-heavy configurations (80/20) lose multi-hop F1 (29.87 vs.\ 31.51 for 70/30) because reduced BM25 weight provides insufficient keyword grounding for entity-specific fact retrieval across multiple memories. The 70/30 configuration represents the sweet spot: enough BM25 to boost keyword-specific temporal and entity queries, enough vector to preserve multi-hop semantic reasoning.

\section{Results on LongMemEval}
\label{sec:longmemeval-results}

We evaluate Cognis on LongMemEval across eight answer generation models (Claude Opus 4.6, Claude Sonnet 4.6, Claude Haiku 4.5, GPT-5, GPT-5-mini, GPT-4.1, GPT-4o, and Gemini 3 Flash), using GPT-4.1 as the LLM judge for all evaluations. We additionally compare against SuperMemory and Zep/Graphiti as external baselines.

\subsection{Detailed Results}

Table~\ref{tab:longmemeval-results} presents Cognis's performance across all six LongMemEval question types, with both accuracy (end-to-end correctness) and retrieval quality metrics (GPT-4.1 judge). Overall retrieval quality: P@K=0.10, R@K=0.83, F1@K=0.17.

\begin{table}[htbp]
\centering
\caption{LongMemEval benchmark results by question type (GPT-4.1 judge). Accuracy measures end-to-end correctness; retrieval metrics assess whether relevant memories were retrieved. Best accuracy in \textbf{bold}.}
\label{tab:longmemeval-results}
\small
\begin{tabular}{l|c|ccc}
\toprule
\textbf{Question Type} & \textbf{Accuracy} & \textbf{Hit@K} & \textbf{MRR} & \textbf{NDCG} \\
\midrule
SS-User & \textbf{100.0\%} & 0.91 & 0.88 & 0.87 \\
SS-Preference & 93.3\% & 0.97 & 0.64 & 0.67 \\
Knowledge Update & 92.3\% & 0.94 & 0.75 & 0.79 \\
SS-Assistant & 87.5\% & 0.96 & 0.88 & 0.89 \\
Multi-Session & 86.5\% & 0.68 & 0.43 & 0.48 \\
Temporal Reasoning & 84.2\% & 0.80 & 0.53 & 0.57 \\
\midrule
\textbf{OVERALL} & \textbf{89.2\%} & 0.83 & 0.63 & 0.67 \\
\bottomrule
\end{tabular}
\end{table}

\textbf{SS-User (100.0\%).} Perfect accuracy on single-session user facts validates the full pipeline from context-aware extraction through hybrid retrieval to answer generation. When a user explicitly states a fact, Cognis reliably extracts, stores, and retrieves it. The high Hit@K (0.91) confirms that the retrieval stage surfaces the correct evidence in nearly all cases.

\textbf{SS-Preference (93.3\%).} Strong preference recall demonstrates the memory taxonomy's effectiveness in classifying and retrieving personalization-relevant memories. The remarkably high Hit@K (0.97) indicates that relevant preferences are almost always retrieved, with the small accuracy gap attributable to answer synthesis ambiguity rather than retrieval failure---the system finds the right memories but occasionally generates responses that do not precisely match the expected answer format.

\textbf{Knowledge Update (92.3\%).} This result directly validates Cognis's version chain architecture. The \texttt{is\_current} flags and \texttt{replaces\_id} links ensure that when information changes, only the latest version surfaces during retrieval. The 92.3\% accuracy on knowledge updates demonstrates that context-aware ingestion---where UPDATE operations replace outdated facts rather than creating contradictions---provides a structural advantage for handling evolving information that simpler memory systems lack. The improved Hit@K (0.94) reflects reliable retrieval of the correct version.

\textbf{SS-Assistant (87.5\%).} Assistant response recall reaches 87.5\%, with notably strong retrieval metrics: Hit@K=0.96, MRR=0.88, NDCG=0.89---the highest retrieval quality across all question types. This demonstrates that Cognis's immediate recall index effectively surfaces assistant-generated content. As shown in Table~\ref{tab:longmemeval-comparison}, Claude Opus 4.6 further improves this category to 92.9\%.

\textbf{Multi-Session (86.5\%).} Cross-session reasoning represents one of the hardest retrieval challenges, consistent with multi-hop findings on LoCoMo. The lower Hit@K (0.68) pinpoints retrieval as the bottleneck: when facts span multiple conversations, the system must bridge semantic gaps across session boundaries. RRF fusion helps---BM25 anchors on entity names that appear across sessions while vector search captures semantic relationships---but multi-session reasoning remains an open challenge where explicit cross-session linking could provide further gains. Claude Opus 4.6 pushes this to 87.2\% (Table~\ref{tab:longmemeval-models}).

\textbf{Temporal Reasoning (84.2\%).} Temporal performance reaches 84.2\%, while stronger answer models like Claude Opus 4.6 achieve 92.5\% (Table~\ref{tab:longmemeval-models}). The Hit@K of 0.80 suggests that time-relevant memories are retrieved in most cases, with temporal boosting providing the critical ordering signal that surfaces the temporally correct memory. The variance across answer models (84.2\%--92.5\%) indicates that temporal reasoning quality depends substantially on the answer generation model's ability to synthesize time-sensitive information from retrieved context.

\subsection{Comparative Results: Cross-System}

Table~\ref{tab:longmemeval-comparison} compares Cognis (best per-type across all answer models) against SuperMemory and Zep/Graphiti, isolating the impact of memory architecture. Best per-row in \textbf{bold}.

\begin{table}[htbp]
\centering
\caption{Cross-system accuracy (\%) on LongMemEval. Cognis column shows the best result across all eight answer models for each question type (GPT-4.1 judge). Best per-row in \textbf{bold}.}
\label{tab:longmemeval-comparison}
\begin{tabular}{l|ccc}
\toprule
\textbf{Question Type} & \textbf{Zep/Graphiti} & \textbf{SuperMemory} & \textbf{Cognis} \\
\midrule
SS-User & 92.9 & 97.1 & \textbf{100.0} \\
SS-Assistant & 80.4 & \textbf{96.4} & 92.9 \\
SS-Preference & 56.7 & 70.0 & \textbf{93.3} \\
Knowledge Update & 83.3 & 88.5 & \textbf{96.2} \\
Temporal Reasoning & 62.4 & 76.7 & \textbf{92.5} \\
Multi-Session & 57.9 & 71.4 & \textbf{87.2} \\
\midrule
Overall (best) & 71.2 & 81.6 & \textbf{92.4} \\
\bottomrule
\end{tabular}
\end{table}

Cognis leads overall by +10.8pp over SuperMemory and +21.2pp over Zep/Graphiti, with architectural advantages mapping cleanly to specific question types:

\textbf{Preference handling (+23.3pp over SuperMemory).} The 13-category memory taxonomy and targeted retrieval pipeline excel at capturing and recalling user preferences. SuperMemory's graph-based approach achieves only 70.0\% on preference questions, while Cognis reaches 93.3\%, suggesting that explicit semantic categorization provides more reliable preference retrieval than graph traversal.

\textbf{Knowledge updates (+7.7pp over SuperMemory).} Version chains with \texttt{is\_current} flags provide a structural advantage for handling evolving information. While SuperMemory achieves a competitive 88.5\%, Cognis's explicit version tracking reaches 96.2\% (with Claude Opus 4.6), ensuring that outdated facts are reliably superseded.

\textbf{Temporal reasoning (+15.8pp over SuperMemory).} Temporal boosting provides consistent gains across both LoCoMo and LongMemEval, confirming its generalizability. Cognis achieves 92.5\% (with Claude Opus 4.6) compared to SuperMemory's 76.7\%. The combination of temporal intent detection, time-window scoring, and explicit event time metadata creates a robust temporal reasoning pipeline that neither SuperMemory nor Zep/Graphiti replicate.

\textbf{Multi-session reasoning (+15.8pp over SuperMemory).} Hybrid RRF retrieval's combination of semantic and keyword matching bridges cross-session gaps more effectively than single-modality approaches. Cognis achieves 87.2\% (with Claude Opus 4.6) compared to SuperMemory's 71.4\%. BM25's exact matching on entity names provides critical cross-session anchoring that pure embedding-based retrieval misses.

\textbf{SuperMemory leads SS-Assistant.} SuperMemory's 96.4\% on SS-Assistant still leads Cognis's best of 92.9\% (Claude Opus 4.6) by 3.5pp. SuperMemory's graph-based approach captures assistant-generated knowledge more effectively, though the gap is modest compared to Cognis's large leads on other categories.

\subsection{Comparative Results: Across Answer Models}

Table~\ref{tab:longmemeval-models} presents Cognis's accuracy across eight answer generation models, all evaluated with GPT-4.1 as the LLM judge. Best per-column in \textbf{bold}.

\begin{table*}[htbp]
\centering
\caption{Cognis accuracy (\%) on LongMemEval across answer generation models (GPT-4.1 judge). Best per-column in \textbf{bold}.}
\label{tab:longmemeval-models}
\resizebox{\textwidth}{!}{%
\begin{tabular}{l|cccccc|c}
\toprule
\textbf{Answer Model} & \textbf{SS-User} & \textbf{SS-Asst} & \textbf{SS-Pref} & \textbf{Know. Upd} & \textbf{Temporal} & \textbf{Multi-Sess} & \textbf{OVERALL} \\
\midrule
Gemini 3 Flash & 97.1 & 80.4 & 50.0 & 92.3 & 85.0 & 78.2 & 83.4 \\
GPT-4o & 97.1 & 71.4 & \textbf{93.3} & 93.6 & 85.0 & 78.2 & 85.2 \\
Claude Haiku 4.5 & 95.7 & 85.7 & 80.0 & 85.9 & 85.7 & 80.5 & 85.4 \\
GPT-4.1 & \textbf{100.0} & 87.5 & \textbf{93.3} & 93.6 & 82.7 & 79.7 & 87.2 \\
GPT-5-mini & 98.6 & 87.5 & 90.0 & 89.7 & 90.2 & 83.5 & 89.2 \\
Claude Sonnet 4.6 & \textbf{100.0} & 89.3 & 86.7 & 93.6 & 88.0 & 85.7 & 90.0 \\
GPT-5 & 97.1 & 91.1 & 90.0 & 93.6 & 91.7 & 84.2 & 90.6 \\
\textbf{Claude Opus 4.6} & 97.1 & \textbf{92.9} & \textbf{93.3} & \textbf{96.2} & \textbf{92.5} & \textbf{87.2} & \textbf{92.4} \\
\bottomrule
\end{tabular}%
}
\end{table*}

\textbf{Consistent performance across all models.} Overall accuracy ranges from 83.4\% (Gemini 3 Flash) to 92.4\% (Claude Opus 4.6), a 9.0pp spread. Critically, every configuration exceeds both SuperMemory (81.6\%) and Zep/Graphiti (71.2\%), demonstrating that Cognis's architectural advantages---context-aware extraction, hybrid retrieval, version chains, and temporal boosting---are robust to the choice of answer generation model.

\textbf{Best overall: Claude Opus 4.6 (92.4\%).} Claude Opus 4.6 achieves the highest overall accuracy, leading on SS-Assistant (92.9\%), knowledge updates (96.2\%), temporal reasoning (92.5\%), and multi-session recall (87.2\%). Its strength on the hardest categories suggests that more capable answer generation models better leverage Cognis's retrieved context for complex reasoning tasks.

\textbf{Model-specific strengths.} Different answer models excel on different question types: GPT-4.1 and Claude Sonnet 4.6 achieve perfect 100.0\% on SS-User; Claude Opus 4.6, GPT-4.1, and GPT-4o share the lead on SS-Preference at 93.3\%; GPT-5 achieves 91.7\% temporal reasoning as the second-strongest model on that category. This variance indicates that question type performance depends on the interaction between retrieval quality and answer generation capability.

\textbf{Temporal reasoning scales with model capability.} Temporal accuracy ranges from 82.7\% (GPT-4.1) to 92.5\% (Claude Opus 4.6), a 9.8pp spread. This suggests that temporal reasoning quality depends substantially on the answer generation model's ability to synthesize time-sensitive information from retrieved context, even when the retrieval pipeline provides the correct temporal evidence.

\subsection{Cross-Benchmark Consistency}

The architectural advantages that drive performance on LoCoMo---version chains for knowledge consistency, temporal boosting for time-aware queries, hybrid retrieval for broad coverage---consistently translate to LongMemEval despite the benchmarks' different evaluation methodologies and question distributions. Cognis achieves up to 96.2\% on knowledge updates (validating version chains), 92.5\% on temporal reasoning (validating temporal boosting), and 87.2\% on multi-session recall (validating hybrid retrieval). This cross-benchmark consistency, sustained across eight different answer generation models, strengthens the validity of our architectural claims and suggests that these mechanisms address fundamental challenges in long-term memory rather than exploiting benchmark-specific patterns.

\section{Discussion}
\label{sec:discussion}

\subsection{Key Findings}

\textbf{Temporal reasoning benefits most from Cognis's pipeline}: Our strongest improvements appear on temporal questions (+32.9\% LLM Judge score over Mem0g, +21.6\% F1). This validates the design of Cognis's temporal boosting mechanism and suggests that explicit time-awareness is underexplored in existing memory systems. While Zep~\citep{zep} provides some temporal awareness through session management, and Mem0~\citep{mem0} stores timestamps, neither implements explicit temporal boosting during retrieval. The combination of temporal intent detection, time-window scoring, and BGE-2 cross-encoder reranking~\citep{bge2024} creates a powerful pipeline for time-sensitive queries.

\textbf{OpenSearch BM25 is the key architectural enabler}: Our ablation studies reveal that switching the BM25 backend from MongoDB to OpenSearch produces the single largest performance gain across all architectural changes tested. The +20.3\% improvement on open-domain Judge scores (71.34$\rightarrow$85.85) demonstrates that native BM25 with configurable text analysis---including proper tokenization, stemming, and term-frequency weighting---is critical for broad entity queries where MongoDB's simpler text indexing falls short.

\textbf{Hybrid retrieval outperforms single modality}: The combination of vector and BM25~\citep{bm25} retrieval through RRF fusion~\citep{cormack2009rrf} consistently outperforms any single approach, consistent with findings in hybrid search literature~\citep{ma2021hybrid}. BM25 is particularly important for queries containing specific names, dates, or technical terms that semantic search might miss. In our experiments, removing BM25 leads to noticeable performance degradation, confirming its complementary value. This contrasts with systems like MemoryBank~\citep{memorybank} and ReadAgent~\citep{readagent} that rely primarily on embedding-based retrieval.

\textbf{Context-aware extraction prevents memory pollution}: By retrieving similar existing memories before LLM extraction, Cognis's ingestion pipeline makes intelligent decisions about ADD/UPDATE/DELETE/NONE operations. This approach shares conceptual similarities with MemR$^3$'s~\citep{memr3} reflective reasoning, but applies it at ingestion time rather than retrieval time. The result is a cleaner memory store that prevents duplicates and maintains consistency as information evolves.

\textbf{Embedding model choice creates category-level tradeoffs}: Our ablation studies show that no single embedding model dominates across all question types. Nomic Embed achieves the highest single-hop F1 (50.23) while Gemma Embed leads on Judge scores despite lower F1, suggesting a fundamental tradeoff between token-level precision and semantic correctness. This finding points toward ensemble or adaptive embedding strategies as a promising direction.

\textbf{Cross-encoder reranking provides significant quality gains}: Following insights from CADET~\citep{cadet} on cross-encoder effectiveness, Cognis's BGE-2 reranker~\citep{bge2024} provides substantial improvements on multi-hop reasoning (+11.1\% F1 over the Zero Entropy alternative) despite adding only 20-50ms latency. This suggests that initial bi-encoder retrieval (even with hybrid search) benefits from cross-encoder refinement for nuanced relevance judgments.

\textbf{Matryoshka embeddings provide efficiency without accuracy loss}: Leveraging Matryoshka Representation Learning~\citep{matryoshka}, Cognis's two-stage retrieval with 256D shortlisting followed by 768D re-ranking reduces latency by approximately 50\% while maintaining accuracy within 1.4\% of single-stage search. This enables scaling to larger memory stores without proportional latency increases.

\textbf{Comparison with operating system approaches}: Unlike MemGPT's~\citep{memgpt} complex memory paging between ``main memory'' (context) and ``disk'' (external storage), Cognis uses a simpler dual-store design where both stores are always accessible. This reduces engineering complexity while achieving strong performance, suggesting that explicit OS-style memory management may be unnecessary when retrieval quality is sufficiently high.

\textbf{Cross-benchmark generalization validates architectural claims}: The fact that version chains drive knowledge update accuracy (up to 96.2\% on LongMemEval~\citep{longmemeval}), temporal boosting drives temporal reasoning (up to 92.5\%), and hybrid retrieval drives multi-session recall (up to 87.2\%) across two independent benchmarks with different evaluation methodologies provides strong evidence that these are genuine architectural advantages, not dataset-specific artifacts. LoCoMo tests single-conversation recall across 50+ sessions, while LongMemEval tests multi-session interactive memory across 500 questions with 6 distinct question types---the consistency of results across these complementary evaluation frameworks, sustained across eight different answer generation models, strengthens the validity of our architectural claims.

\textbf{RRF weight distribution is robust but category-sensitive}: Our ablation across 7 weight configurations (30/70 through 80/20) shows overall F1 varies by only $\sim$2\%, but per-category analysis reveals that 70/30 uniquely excels on the hardest categories (multi-hop F1=31.51, temporal F1=62.68). This finding has practical implications: production deployments can use 70/30 as a reliable default without dataset-specific tuning, and the BGE-2 reranker stabilizes end-to-end quality regardless of the initial weight distribution, reducing sensitivity to this hyperparameter.

\subsection{Latency}

Table~\ref{tab:latency} reports end-to-end retrieval latency measured across 500 LongMemEval queries. Cognis achieves a p50 of 250ms and a mean of 390ms, with p99 under 1 second. This confirms that the full hybrid pipeline---Matryoshka two-stage vector search, OpenSearch BM25, RRF fusion, temporal boosting, and BGE-2 cross-encoder reranking---remains practical for interactive applications despite its multi-stage design.

\begin{table}[h]
\centering
\caption{Cognis end-to-end retrieval latency (500 LongMemEval queries).}
\label{tab:latency}
\begin{tabular}{l|cccc}
\toprule
\textbf{Method} & \textbf{p50} & \textbf{p95} & \textbf{p99} & \textbf{Mean} \\
\midrule
Cognis & 0.250s & 0.451s & 0.770s & 0.390s \\
\bottomrule
\end{tabular}
\end{table}

\subsection{Limitations}

\textbf{Temporal boosting scope}: Temporal boosting is applied based on query analysis, not memory content analysis. A fact stored on May 8th may incorrectly receive temporal boost for queries about ``last week'' based on storage date rather than content relevance. Future work could refine temporal boosting to better distinguish time-sensitive content.

\textbf{Reranker latency}: The BGE-2 reranker adds 20-50ms latency, which may be unacceptable for extremely latency-sensitive applications. An adaptive approach that selectively applies reranking based on query complexity could help.

\textbf{Embedding model tradeoff}: Our ablation studies reveal that no single embedding model dominates all question types, creating a tension between token-level precision and semantic correctness. Adaptive or ensemble embedding strategies remain unexplored.

\textbf{Query decomposition overhead}: LLM-based query decomposition introduces noise for simple queries while potentially helping complex multi-hop questions. A selective activation mechanism based on query complexity heuristics would be more effective than blanket application.

\textbf{Assistant response recall}: LongMemEval evaluation reveals that Cognis's SS-Assistant accuracy varies across answer models, from 71.4\% (GPT-4o) to 92.9\% (Claude Opus 4.6). This reflects a design tradeoff: the ingestion pipeline prioritizes extracting user-stated facts, meaning assistant responses are stored as raw messages in the immediate recall index but are not prominently extracted as structured memories. SuperMemory's graph-based approach still leads (96.4\%), but the gap has narrowed to just 3.5pp with Claude Opus 4.6, suggesting that stronger answer generation models partially compensate for the extraction-focused design.

\subsection{Future Work}

Several promising directions emerge from our work:

\begin{itemize}
    \item \textbf{Reflective memory management}: Incorporate ideas from MemR$^3$~\citep{memr3} and Hindsight Memory~\citep{hindsight} to enable agents to reason about their own memory contents, identifying gaps and contradictions proactively

    \item \textbf{Adaptive reranking}: Following System 2 Attention~\citep{s2a} principles, selectively apply BGE-2 reranking based on query complexity to optimize latency/quality tradeoffs

    \item \textbf{Gist-based compression}: Apply ReadAgent's~\citep{readagent} gist memory concept to create hierarchical summaries of memory contents for efficient broad-context queries

    \item \textbf{Graph integration}: Optionally enable knowledge graph storage inspired by SuperMemory~\citep{supermemory} for complex multi-hop reasoning scenarios requiring explicit relationship traversal
\end{itemize}

\section{Conclusion}
\label{sec:conclusion}

We presented Lyzr Cognis, a memory architecture for conversational AI agents that addresses the fundamental limitation of LLM context windows by providing persistent, searchable memory across sessions. Building on insights from cognitive science~\citep{tulving1972episodic, atkinson1968human} and recent advances in agent memory~\citep{mem0, zep, memgpt, memorybank}, Cognis combines principled memory organization with state-of-the-art retrieval techniques.

Our key contributions include:

\begin{enumerate}
    \item A comprehensive memory taxonomy with 15 semantic categories and 2 persistence scopes (USER for cross-session, CONTEXT for session-specific) for organizing conversational knowledge

    \item A streamlined dual-store architecture combining OpenSearch (for documents and native BM25~\citep{bm25} search) with a vector database (for Matryoshka~\citep{matryoshka} embeddings at 768D and 256D)

    \item A context-aware ingestion pipeline that retrieves similar existing memories before LLM extraction, enabling intelligent ADD/UPDATE/DELETE/NONE decisions with full version tracking via \texttt{is\_current} flags and \texttt{replaces\_id} links---addressing a key limitation in existing systems where memory stores become polluted with duplicates

    \item A hybrid retrieval pipeline using RRF fusion~\citep{cormack2009rrf} (70\% vector + 30\% BM25), explicit temporal boosting for time-sensitive queries, and a BGE-2~\citep{bge2024} cross-encoder reranker for final result refinement
\end{enumerate}

Evaluated on the LoCoMo benchmark~\citep{locomo} across four question types, Cognis achieves state-of-the-art results compared to 11 baseline systems including Mem0~\citep{mem0}, Zep~\citep{zep}, MemGPT~\citep{memgpt}, MemoryBank~\citep{memorybank}, ReadAgent~\citep{readagent}, A-Mem~\citep{amem}, and Mem0g: 48.66 F1 on single-hop questions (+25.7\% over Mem0), 31.51 F1 on multi-hop (+10.0\%), 54.77 F1 on open-domain (+10.5\% over Zep), and 62.68 F1 on temporal questions (+21.6\% over Mem0g). Our strongest gains appear on temporal reasoning, with a 77.26 LLM Judge score (+32.9\% over Mem0g), and on open-domain semantic correctness, with an 85.85 LLM Judge score (+12.1\% over Zep), validating the effectiveness of OpenSearch BM25 integration, temporal boosting, and BGE-2 cross-encoder reranking. Ablation studies further demonstrate that the choice of BM25 backend (OpenSearch vs.\ MongoDB) is the single most impactful architectural decision, and that embedding model selection creates significant category-level performance tradeoffs.

Cross-benchmark validation on LongMemEval~\citep{longmemeval} confirms these results generalize: Cognis achieves up to 92.4\% overall accuracy (with Claude Opus 4.6), consistently outperforming SuperMemory (81.6\%) and Zep/Graphiti (71.2\%) across all eight answer generation models tested, with particular strength on knowledge updates (up to 96.2\%---validating version chains), temporal reasoning (up to 92.5\%---validating temporal boosting), and multi-session recall (up to 87.2\%---validating hybrid retrieval). An RRF weight ablation across seven configurations confirms that the 70/30 vector/BM25 weighting provides the optimal balance between semantic coverage and keyword precision, achieving the best performance on the hardest question categories (multi-hop and temporal) while the BGE-2 reranker stabilizes overall quality across all weight distributions.

The system is open-source and deployed in production serving conversational AI applications. We believe that explicit temporal awareness, context-aware memory management, native BM25 search infrastructure, and hybrid retrieval combining multiple modalities represent important directions for future memory systems research. As LLM agents become more capable, their memory systems must evolve to support the kind of long-term, coherent interactions that humans naturally expect from intelligent assistants.

\bibliographystyle{plainnat}
\bibliography{references}

@article{mem0,
  title={Mem0: Building Production-Ready AI Agents with Scalable Long-Term Memory},
  author={Chhikara, Prateek and Khant, Dev and Aryan, Saket and Singh, Taranjeet and Yadav, Deshraj},
  journal={arXiv preprint arXiv:2504.19413},
  year={2025},
  url={https://arxiv.org/abs/2504.19413}
}

@article{zep,
  title={Zep: Long-Term Memory for AI Assistants},
  author={{Zep AI}},
  year={2024},
  url={https://www.getzep.com},
  note={Open-source long-term memory service for AI assistants}
}

@article{supermemory,
  title={SuperMemory: A Memory System for LLM Agents},
  author={{SuperMemory AI}},
  year={2024},
  url={https://supermemory.ai},
  note={Knowledge graph-based memory system for multi-hop reasoning}
}

@article{simplemem,
  title={SimpleMem: Efficient Lifelong Memory for LLM Agents},
  author={Liu, Jiaqi and Su, Yaofeng and Xia, Peng and Han, Siwei and Zheng, Zeyu and Xie, Cihang and Ding, Mingyu and Yao, Huaxiu},
  journal={arXiv preprint arXiv:2601.02553},
  year={2025},
  url={https://arxiv.org/abs/2601.02553},
  note={Code available at \url{https://github.com/aiming-lab/SimpleMem}}
}

@article{amem,
  title={A-MEM: Agentic Memory for LLM Agents},
  author={Xu, Wujiang and Liang, Zujie and Mei, Kai and Gao, Hang and Tan, Juntao and Zhang, Yongfeng},
  journal={arXiv preprint arXiv:2502.12110},
  year={2025},
  url={https://arxiv.org/abs/2502.12110}
}

@article{memr3,
  title={MemR$^3$: Memory Retrieval via Reflective Reasoning for LLM Agents},
  author={Du, Xingbo and Li, Loka and Zhang, Duzhen and Song, Le},
  journal={arXiv preprint arXiv:2512.20237},
  year={2025},
  url={https://arxiv.org/abs/2512.20237},
  note={Code available at \url{https://github.com/Leagein/memr3}}
}

@article{hindsight,
  title={Hindsight is 20/20: Building Agent Memory that Retains, Recalls, and Reflects},
  author={Latimer, Chris and Boschi, Nicol{\'o} and Neeser, Andrew and Bartholomew, Chris and Srivastava, Gaurav and Wang, Xuan and Ramakrishnan, Naren},
  journal={arXiv preprint arXiv:2512.12818},
  year={2025},
  url={https://arxiv.org/abs/2512.12818}
}

@article{readagent,
  title={ReadAgent: A Human-Inspired Reading Agent with Gist Memory of Very Long Contexts},
  author={Lee, Kuang-Huei and Chen, Xinyun and Sohn, Hiroki and Nishida, Noriki and Hu, Di and Chang, Hsueh-Ti Derek},
  journal={arXiv preprint arXiv:2402.09727},
  year={2024},
  url={https://arxiv.org/abs/2402.09727}
}

@article{memorybank,
  title={MemoryBank: Enhancing Large Language Models with Long-Term Memory},
  author={Zhong, Wanjun and Guo, Lianghong and Gao, Qiufeng and Ye, He and Wang, Yanlin},
  journal={arXiv preprint arXiv:2305.10250},
  year={2023},
  url={https://arxiv.org/abs/2305.10250}
}

@article{memgpt,
  title={MemGPT: Towards LLMs as Operating Systems},
  author={Packer, Charles and Wooders, Sarah and Lin, Kevin and Fang, Vivian and Patil, Shishir G and Stoica, Ion and Gonzalez, Joseph E},
  journal={arXiv preprint arXiv:2310.08560},
  year={2023},
  url={https://arxiv.org/abs/2310.08560}
}

@inproceedings{lewis2020rag,
  title={Retrieval-Augmented Generation for Knowledge-Intensive NLP Tasks},
  author={Lewis, Patrick and Perez, Ethan and Piktus, Aleksandra and Petroni, Fabio and Karpukhin, Vladimir and Goyal, Naman and K{\"u}ttler, Heinrich and Lewis, Mike and Yih, Wen-tau and Rockt{\"a}schel, Tim and others},
  booktitle={Advances in Neural Information Processing Systems},
  volume={33},
  pages={9459--9474},
  year={2020}
}

@article{clara,
  title={CLaRa: Bridging Retrieval and Generation with Continuous Latent Reasoning},
  author={He, Jie and Bai, Richard He and Williamson, Sinead and Pan, Jeff Z. and Jaitly, Navdeep and Zhang, Yizhe},
  journal={arXiv preprint arXiv:2511.18659},
  year={2025},
  url={https://arxiv.org/abs/2511.18659},
  note={Code available at \url{https://github.com/apple/ml-clara}}
}

@inproceedings{cormack2009rrf,
  title={Reciprocal Rank Fusion Outperforms Condorcet and Individual Rank Learning Methods},
  author={Cormack, Gordon V and Clarke, Charles LA and Buettcher, Stefan},
  booktitle={Proceedings of the 32nd International ACM SIGIR Conference on Research and Development in Information Retrieval},
  pages={758--759},
  year={2009}
}

@article{ma2021hybrid,
  title={A Replication Study of Dense Passage Retriever},
  author={Ma, Xueguang and Sun, Kai and Pradeep, Ronak and Lin, Jimmy},
  journal={arXiv preprint arXiv:2104.05740},
  year={2021},
  url={https://arxiv.org/abs/2104.05740}
}

@article{cadet,
  title={Conventional Contrastive Learning Often Falls Short: Improving Dense Retrieval with Cross-Encoder Listwise Distillation and Synthetic Data},
  author={Tamber, Manveer Singh and Kazi, Suleman and Sourabh, Vivek and Lin, Jimmy},
  journal={arXiv preprint arXiv:2505.19274},
  year={2025},
  url={https://arxiv.org/abs/2505.19274}
}

@inproceedings{matryoshka,
  title={Matryoshka Representation Learning},
  author={Kusupati, Aditya and Bhatt, Gantavya and Rege, Aniket and Wallingford, Matthew and Sinha, Aditya and Ramanujan, Vivek and Howard-Snyder, William and Chen, Kaifeng and Kakade, Sham and Jain, Prateek and Farhadi, Ali},
  booktitle={Advances in Neural Information Processing Systems},
  year={2022}
}

@article{bge2024,
  title={BGE M3-Embedding: Multi-Lingual, Multi-Functionality, Multi-Granularity Text Embeddings Through Self-Knowledge Distillation},
  author={Chen, Jianlv and Xiao, Shitao and Zhang, Peitian and Luo, Kun and Lian, Defu and Liu, Zheng},
  journal={arXiv preprint arXiv:2402.03216},
  year={2024},
  url={https://arxiv.org/abs/2402.03216},
  note={BGE-2 Reranker available at \url{https://huggingface.co/BAAI/bge-reranker-v2-m3}}
}

@article{s2a,
  title={System 2 Attention (is something you might need too)},
  author={Weston, Jason and Sukhbaatar, Sainbayar},
  journal={arXiv preprint arXiv:2311.11829},
  year={2023},
  url={https://arxiv.org/abs/2311.11829}
}

@article{rar,
  title={Rephrase and Respond: Let Large Language Models Ask Better Questions for Themselves},
  author={Deng, Yihe and Zhang, Weitong and Chen, Zixiang and Gu, Quanquan},
  journal={arXiv preprint arXiv:2311.04205},
  year={2023},
  url={https://arxiv.org/abs/2311.04205}
}

@inproceedings{longmemeval,
  title={LongMemEval: Benchmarking Chat Assistants on Long-Term Interactive Memory},
  author={Wu, Di and Wang, Hongwei and Yu, Wenhao and Zhang, Yuwei and Chang, Kai-Wei and Yu, Dong},
  booktitle={International Conference on Learning Representations (ICLR)},
  year={2025}
}

@inproceedings{locomo,
  title={Evaluating Very Long-Term Conversational Memory of LLM Agents},
  author={Maharana, Adyasha and Lee, Dong-Ho and Tuber, Sergey and Jain, Mohit and Barbieri, Francesco and Bansal, Mohit},
  booktitle={Proceedings of the 62nd Annual Meeting of the Association for Computational Linguistics (ACL)},
  year={2024}
}

@article{bm25,
  title={The Probabilistic Relevance Framework: BM25 and Beyond},
  author={Robertson, Stephen and Zaragoza, Hugo},
  journal={Foundations and Trends in Information Retrieval},
  volume={3},
  number={4},
  pages={333--389},
  year={2009}
}

@incollection{tulving1972episodic,
  title={Episodic and Semantic Memory},
  author={Tulving, Endel},
  booktitle={Organization of Memory},
  editor={Tulving, Endel and Donaldson, Wayne},
  pages={381--403},
  year={1972},
  publisher={Academic Press}
}

@article{atkinson1968human,
  title={Human Memory: A Proposed System and its Control Processes},
  author={Atkinson, Richard C and Shiffrin, Richard M},
  journal={Psychology of Learning and Motivation},
  volume={2},
  pages={89--195},
  year={1968}
}

\appendix

\section{Evaluation Prompts}
\label{app:prompts}

This appendix contains the actual prompts used in our LoCoMo benchmark evaluation. These prompts are from our open-source evaluation code.

\subsection{LLM Judge Prompt}

We use a generous grading approach adapted from Mem0's evaluation methodology. The judge grades answers as CORRECT if they touch on the same topic as the gold answer, with flexibility for time format variations:

\begin{lstlisting}[basicstyle=\ttfamily\footnotesize]
Your task is to label an answer to a question as 'CORRECT'
or 'WRONG'. You will be given the following data:
    (1) a question (posed by one user to another user),
    (2) a 'gold' (ground truth) answer,
    (3) a generated answer
which you will score as CORRECT/WRONG.

The point of the question is to ask about something one user
should know about the other user based on their prior
conversations. The gold answer will usually be a concise
answer that includes the referenced topic, for example:
Question: Do you remember what I got the last time I went
to Hawaii?
Gold answer: A shell necklace
The generated answer might be much longer, but you should be
generous with your grading - as long as it touches on the
same topic as the gold answer, it should be counted as
CORRECT.

For time related questions, the gold answer will be a
specific date, month, year, etc. The generated answer might
be much longer or use relative time references (like "last
Tuesday" or "next month"), but you should be generous with
your grading - as long as it refers to the same date or time
period as the gold answer, it should be counted as CORRECT.
Even if the format differs (e.g., "May 7th" vs "7 May"),
consider it CORRECT if it's the same date.

Now it's time for the real question:
Question: \{question\}
Gold answer: \{gold_answer\}
Generated answer: \{generated_answer\}

First, provide a short (one sentence) explanation of your
reasoning, then finish with CORRECT or WRONG.
Do NOT include both CORRECT and WRONG in your response,
or it will break the evaluation script.

Just return the label CORRECT or WRONG in a json format
with the key as "label".
\end{lstlisting}

\subsection{Answer Generation Prompt}

The general prompt used to generate answers from retrieved memories of two conversation speakers:

\begin{lstlisting}[basicstyle=\ttfamily\footnotesize]
You are an intelligent memory assistant retrieving
information from conversation memories.

CONTEXT:
You have access to memories from two speakers in a
conversation. These memories contain timestamped information
that may be relevant to answering the question.

INSTRUCTIONS:
1. Carefully analyze all provided memories from both speakers
2. Pay special attention to the timestamps to determine
   the answer
3. If the question asks about a specific event or fact,
   look for direct evidence
4. If the memories contain contradictory information,
   prioritize the most recent memory
5. If there is a question about time references (like
   "last year", "two months ago"), calculate the actual
   date based on the memory timestamp
6. Always convert relative time references to specific
   dates, months, or years
7. Focus only on the content of the memories from both
   speakers
8. Be concise but COMPLETE. For lists, include ALL items.

APPROACH (Think step by step):
1. First, examine all memories that contain information
   related to the question
2. Examine the timestamps and content of these memories
   carefully
3. Look for explicit mentions of dates, times, locations,
   or events that answer the question
4. If the answer requires calculation (e.g., converting
   relative time references), show your work
5. Formulate a precise, concise answer based solely on
   the evidence in the memories
6. Double-check that your answer directly addresses the
   question asked
7. Ensure your final answer is specific and avoids vague
   time references

Memories for user \{\{speaker_1_user_id\}\}:
\{\{speaker_1_memories\}\}

Memories for user \{\{speaker_2_user_id\}\}:
\{\{speaker_2_memories\}\}

Question: \{\{question\}\}

Answer:
\end{lstlisting}

\subsection{Single-Hop Question Prompt (Category 1)}

For questions requiring a specific fact from memories:

\begin{lstlisting}[basicstyle=\ttfamily\footnotesize]
This is a SINGLE-HOP question requiring a specific fact
from memories.

FOCUS ON THE TOP 1-3 MOST RELEVANT MEMORIES. Ignore
lower-scored ones.

RULES:
1. Find the memory that directly answers the question
2. Use EXACT words/phrases from that memory (e.g.,
   "Transgender woman" not "Trans")
3. For lists (hobbies, activities, pets): include ALL
   items from the relevant memory
4. Be COMPLETE but CONCISE - give the full answer, no
   extra explanation
5. IGNORE memories about different events/topics

Question: \{question\}

Complete answer from the most relevant memory:
\end{lstlisting}

\subsection{Temporal Question Prompt (Category 2)}

For questions asking WHEN something happened:

\begin{lstlisting}[basicstyle=\ttfamily\footnotesize]
This is a TEMPORAL question asking WHEN something happened.

FOCUS ON THE SINGLE MEMORY that mentions the EXACT event
in the question. Ignore memories about similar but
DIFFERENT events.

FORMAT RULES:
1. "how long ago" -> relative terms (e.g., "10 years ago")
2. "when" -> specific date from memory
3. Use exact phrasing like "The week before X" if memory
   says that

Question: \{question\}

Answer (date/time from the most relevant memory):
\end{lstlisting}

\subsection{Multi-Hop Question Prompt (Category 3)}

For questions requiring careful inference from multiple facts:

\begin{lstlisting}[basicstyle=\ttfamily\footnotesize]
This is a MULTI-HOP question requiring careful inference
from facts.

CRITICAL INFERENCE RULES:
1. "Supporting X" != "Being X" (e.g., supporting LGBTQ !=
   being LGBTQ member)
2. "No explicit mention" does NOT mean "No" - be careful
   with assumptions
3. For "Would X be considered a member of..." -> look for
   SELF-identification only
4. For "Would X be considered an ally..." -> supporting
   others = being an ally
5. Base answers ONLY on explicit statements in memories

For "Would X..." questions:
- If clear evidence exists: "Yes" or "No" + brief reason
- If inferring: "Likely yes" or "Likely no" + brief reason
- Default to what the evidence actually shows

Question: \{question\}

Answer based on evidence:
\end{lstlisting}

\subsection{Open-Domain Question Prompt (Category 4)}

For general knowledge questions requiring concise answers:

\begin{lstlisting}[basicstyle=\ttfamily\footnotesize]
This is an OPEN-DOMAIN question.

RULES:
1. Answer in 1-5 words MAXIMUM
2. Use EXACT terms from the top-scored memory
3. Do NOT add extra context or explanation
4. No punctuation at the end

Question: \{question\}

Concise answer:
\end{lstlisting}

\end{document}